%% file: main.tex
\documentclass[journal]{vgtc}                     

\onlineid{1394}
\vgtccategory{Research}

\newcommand{\ourmodel}{Chart2SVG\xspace}
\title{\ourmodel: Editable SVG Generation from Raster Chart Images} 

\author{
  \authororcid{Jinning Cui}{},
\authororcid{Lu Chen}{0000-0002-2628-3876},
\authororcid{Haoyan Shi}{}, 
\authororcid{Yue He}{}, 
\authororcid{Chenglong Wang}{}, 
\authororcid{Mengyu Zhou}{}, \\
\authororcid{Weidong Huang}{}, and
\authororcid{Yunhai Wang}{0000-0003-0059-6580}
}

\authorfooter{
  \item  Jinning Cui, Yue He and Yunhai Wang are with Renmin University of China, Beijing, China.
    Email: \{jinningcui,heyue25,wang.yh\}@ruc.edu.cn.
\item Lu Chen is with State Key Lab of CAD\&CG, Zhejiang University, Zhejiang, China. 
    Email: lu.chen@zju.edu.cn
 \item Haoyan Shi is with Shandong University, Shandong, China. 
    Email: 202315173@mail.sdu.edu.cn
 \item Chenglong Wang is with Microsoft Research, Redmond, Washington, United States.
    Email: chenglong.wang@microsoft.com
  \item Mengyu Zhou is with Qwen Large Model Application Team, Alibaba, Beijing, China.
    Email: zhoumengyu.zmy@alibaba-inc.com
  \item Weidong Huang is with University of Technology Sydney, Sydney, NSW, Australia.
    Email: Weidong.Huang@uts.edu.au
  \item Yunhai Wang is the corresponding author.  
}

\abstract{%
  \input{sections/0_abstract}
}

\keywords{Chart Reverse Engineering, Scalable Vector Graphics, Vision-Language Models}

\teaser{
  \centering
  \includegraphics[width=\linewidth, alt={A view of a city with buildings peeking out of the clouds.}]{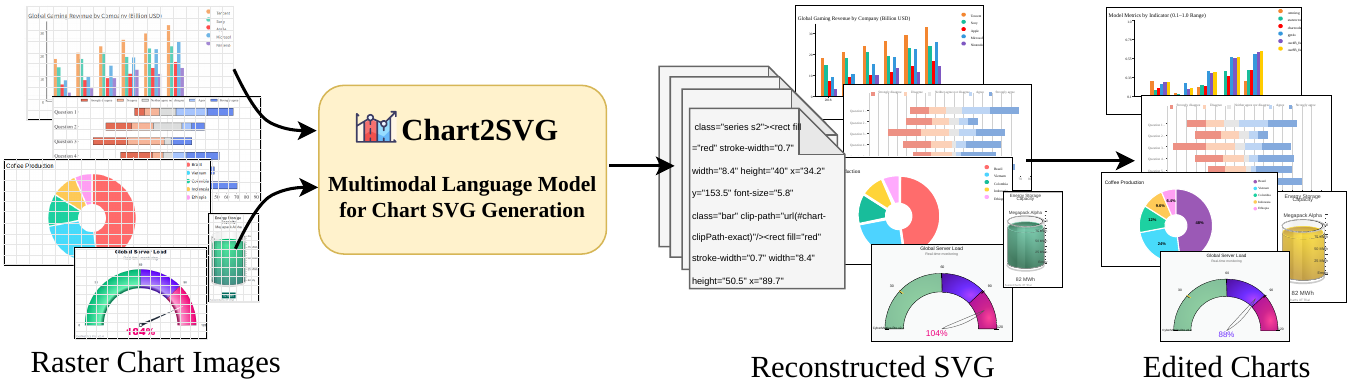}
\caption{%
\ourmodel processes raster chart images using a multimodal LLM architecture and converts a wide range of visualizations, including bar charts, pie charts, and custom designs, into high-quality Scalable Vector Graphics (SVG) code. This enables downstream tasks such as data exploration, chart repurposing, and layout reuse. (Left) Inputs: diverse raster chart images. (Center) \new{Our multimodal language model trained on large-scale chart corpus}. (Right) Outputs: generated \new{SVG source code that represents semantically structured and editable vectorized charts that support language-guided manipulation for downstream tasks}.
}
  \label{fig:teaser}
}

\graphicspath{{figs/}{figures/}{pictures/}{images/}{./}} 

\usepackage{tabu}                      
\usepackage{booktabs}                  
\usepackage{lipsum}                    
\usepackage{mwe}                       
\usepackage{ccicons}                   
\usepackage{pifont}
\usepackage{amsmath}
\usepackage{algorithm}
\usepackage{algpseudocode}
\usepackage{multirow}
\usepackage{booktabs}
\usepackage{tabularx}
\usepackage{array}
\usepackage{tcolorbox}
\usepackage{booktabs}
\usepackage[table]{xcolor}
\usepackage{graphicx}
\usepackage{xcolor}
\definecolor{newcolor}{RGB}{0, 0, 0}  
\newcommand{\new}[1]{\textcolor{newcolor}{#1}}

\usepackage[pagebackref,bookmarks]{hyperref}
\hypersetup{
    colorlinks=true,
    linkcolor=blue,
    urlcolor=blue,
    citecolor=blue
}

\usepackage{mathptmx}                  

\newcommand{\myparagraph}[1]{\mbox{\ } \newline \noindent \textbf{#1}}
\renewcommand{\paragraph}[1]{\vspace{-2.5mm}\myparagraph{#1}}

\begin{document}


\maketitle
\input{sections/1_intro_refine}
\input{sections/2_related}

\input{sections/3_method}
\input{sections/4_application}

\input{sections/5_evaluation}
\input{sections/6_case-study}

\input{sections/6_disscusion}
\input{sections/7_Ack.tex}

\bibliographystyle{abbrv-doi-hyperref}

\bibliography{chartsvg}

\clearpage
\input{appendix}

\end{document}

%% file: sections/0_abstract.tex
We present \ourmodel, a multimodal large language model that converts static raster charts into \new{structurally organized, semantically enriched SVGs that support programmatic editing}. By incorporating chart-specific semantic tokens into a vision-language model, \ourmodel captures both geometric primitives and their functional roles. To support robust structural recovery, we introduce Beagle+, a dataset of 33K canonicalized and structurally distilled chart samples. Our approach combines specialized training objectives with a rendering-aware post-training phase, producing SVGs that are both visually accurate and structurally consistent. To facilitate higher-level manipulations, we construct a Chart Structure Graph (CSG) that exposes visual dependencies, enabling tasks such as interactive exploration, chart repurposing, and layout reuse. Experiments show that \ourmodel substantially outperforms baselines in reconstruction fidelity and \new{downstream editing utility}, advancing the development of intelligent and interactive visualization tools.
The code is available at \url{https://github.com/JinningCui/Chart2SVG}. 
The dataset is available at \url{https://huggingface.co/datasets/syslocker/Beagle_Plus}.

%% file: sections/1_intro_refine.tex
\section{Introduction}
Data visualizations, such as charts, are a cornerstone of communication, distilling complex insights across scientific research, business analytics, and journalism. To facilitate broad knowledge dissemination, these visualizations are often rendered and shared as static raster images. However, this flattening process removes their underlying structure and data relationships, making reuse or modification notoriously difficult. Consequently, many advanced reuse techniques focus exclusively on scalable vector graphics (SVGs)~\cite{cui2021mixed,chen2023mystique,snyder2023divi}. At the same time, charts frequently need to be adapted to meet evolving requirements, from localized updates, such as correcting a specific data value, to global stylistic adjustments, such as refining color palettes for accessibility. This continual refinement is central to the reuse and modification process, ensuring charts remain accurate, effective, and communicatively relevant as their context evolves~\cite{bigelow2016iterating,bako2022understanding}.

Traditionally, reusing and modifying high-fidelity rasterized charts has relied on labor-intensive manual workflows, typically requiring users to convert images into SVGs using professional tools like Adobe Illustrator~\cite{harder2023creating}, followed by extensive cleanup. To bypass these bottlenecks, a substantial body of work has explored reverse-engineering visualizations directly from bitmap images to recover underlying data and visual encodings, rather than reconstructing the original SVG. Early systems, such as ReVision~\cite{savva2011revision}, pioneered automated chart classification and graphical mark extraction. Subsequent approaches refined these pipelines for greater reliability. For example, Poco and Heer~\cite{poco2017reverse} focused on extracting precise encoding mappings, while ChartSense~\cite{jung2017chartsense} introduced interactive, human-in-the-loop workflows. Deep learning methods, such as ChartOCR~\cite{luo2021chartocr}, further advanced the field by leveraging object detection and optical character recognition (OCR) to robustly reconstruct data tables and chart structures.

In parallel, another line of research in the machine learning community explores direct vectorization, converting raster images into SVGs. Early deep learning approaches, such as Im2Vec~\cite{reddy2021im2vec} and LIVE~\cite{ma2022towards}, treat charts as generic graphics, capturing visual paths but losing the semantic binding between marks and underlying data. The emergence of multimodal large language models (MLLMs)~\cite{yin2024survey} enables end-to-end parsing, generating structured representations of data and visual encodings without separate extraction steps. For example, StarVector~\cite{rodriguez2025starvector} outputs compilable SVG syntax, while OneChart~\cite{chen2024onechart} improves numerical reasoning using auxiliary tokens. Chart-to-code methods, such as ChartCoder~\cite{zhao2025chartcoder}, extend this paradigm by generating executable visualization code directly from images. Despite these advances, none of these methods directly recover the original chart SVG structure, so precise graphical elements, styling, and spatial relationships are often lost. As a result, automated reproduction or programmatic modification may fail to fully preserve the chart’s visual fidelity and structure.

To address these challenges, we introduce \ourmodel, a foundational MLLM for chart SVG generation and editing. \ourmodel leverages a vision-language model (VLM) adapted from Qwen3-VL~\cite{bai2025qwen3} to reconstruct structural SVG charts from raster inputs (as shown in Figure~\ref{fig:teaser}), capturing both geometric primitives and their underlying semantics. To mitigate the stochasticity in raw vector data, we introduce a chart SVG normalization pipeline that canonicalizes geometry, coordinate systems, and styling, improving training stability and reducing variability. During supervised fine-tuning, the model is augmented with SVG semantic tokens to explicitly encode chart components and their relationships, enabling more structured and accurate generation. The model is further optimized using rendering-aware Group Relative Policy Optimization (GRPO)~\cite{shao2024deepseekmath, rodriguez2025rendering}, which incorporates execution-based visual feedback to penalize structural inconsistencies and invalid code. This ensures outputs are both visually faithful and programmatically correct, providing a stable foundation for downstream editing.

Once the SVG is reconstructed, we apply heuristics from~\cite{snyder2023divi,chen2023mystique} to \new{label code primitives as semantic components} such as axes, legends, titles, and marks.
\new{Building on these labels, we introduce a Chart Structure Graph (CSG), which lifts the primitive-level SVG annotations into a structured edit-propagation graph. The CSG captures dependencies in data encoding and spatial layout among primitives, providing a grounded reasoning substrate for an LLM to plan coordinated edits across spatially scattered yet inherently connected SVG elements.
This enables complex chart-editing tasks, such as recoloring data points (while updating the legend) or adjusting axes (while repositioning marks), to be performed automatically while preserving structural consistency} and visual integrity. By combining execution-guided SVG recovery with graph-based semantic reasoning, \ourmodel bridges high-level user intent and low-level vector manipulation, enabling robust, programmatic editing of complex charts.

We evaluate \ourmodel along two dimensions: reconstruction quality and downstream utility. 
For reconstruction, we compare against representative baselines on held-out in-distribution benchmarks covering multiple chart-generation toolchains, as well as an out-of-distribution benchmark of real-world charts. 
Our evaluation measures not only pixel fidelity and perceptual similarity, but also foreground alignment, edge consistency, and rendering failure rate, reflecting both visual quality and executability. 
We further conduct ablations to isolate the effects of chart-specific semantic tokens and rendering-aware GRPO. 
Beyond reconstruction, we use case studies to examine the practical value of the recovered representation, showing that \ourmodel supports language-guided chart analysis, semantic editing, chart reuse with new data, and multi-chart composition.

In summary, our contributions are threefold:
\begin{itemize}
    \item \new{We present \ourmodel, a MLLM for reconstructing 
    editable SVG charts from raster images, trained on Beagle+, a 33K-sample canonicalized chart SVG corpus designed to regularize heterogeneous web-crawled SVGs into a more learnable representation;}
    \item \new{We introduce Chart Structure Graph, an LLM-facing interface for structure-aware direct manipulation that lifts reconstructed SVG primitives into a semantic dependency space for propagating edits automatically and consistently across related SVG elements;}
    \item \new{We evaluate \ourmodel and Chart Structure Graph through real-world reconstruction benchmarks, ablation studies, and downstream editing tasks, demonstrating strong reconstruction fidelity and practical editing utility.}
\end{itemize}

%% file: sections/2_related.tex
\section{Related Work}

\subsection{Reverse-Engineering Visualizations}
Prior work on chart reverse engineering has primarily focused on recovering chart structures, visual encodings, or underlying data from static rendered images. Early systems, such as ReVision~\cite{savva2011revision}, treated chart images as bitmap inputs for classification and mark extraction, demonstrating that visualizations can be partially repurposed even without access to the original data. Similarly, Poco and Heer~\cite{poco2017reverse} targeted the recovery of Vega-Lite representations, using a pipeline of OCR and mark-type recognition to systematically infer data domains and encoding ranges. To improve the accuracy of fully automated pipelines, ChartSense~\cite{jung2017chartsense} introduced an interactive, human-in-the-loop workflow, enabling users to refine extracted elements and achieve precise data recovery. Beyond general reuse, these techniques are also critical for accessibility; for example, Choi et al.~\cite{choi2019visualizing} demonstrated that recovering structured data from static charts enables alternative modalities, such as tactile graphics or sonified descriptions, for visually impaired users.

Subsequent work strengthened these pipelines with learned detectors and multimodal reasoning. 
Subsequent work strengthened these pipelines with learned detectors: ChartOCR~\cite{luo2021chartocr} predicts keypoints to reconstruct data with chart-specific rules; LineEX~\cite{hassan2023lineex} recovers lines and scales via pose estimation; and ChartReader~\cite{cheng2023chartreader} integrates transformer-based detection with pretrained VLMs for chart-to-table and question answering tasks.
More recently, MLLMs~\cite{wang2025llm4dsr} have enabled end-to-end generative reconstruction via code. OneChart~\cite{chen2024onechart} parses data and encodings in a single pass using structured tokens and auxiliary numerical grounding tokens. Benchmarks such as Plot2Code~\cite{wu2025plot2code} and ChartX~\cite{xia2025chartx} evaluate MLLMs on their ability to reproduce chart data and textual content, while ChartCoder~\cite{zhao2025chartcoder} generates executable visualization code directly from images. Despite these advances, such methods often fail to preserve the precise spatial and structural fidelity of the original chart, producing non-renderable outputs or distorted geometries.

Despite significant progress, existing methods typically recover tables or specifications rather than editable vector graphics, leaving visually faithful and reusable SVG reconstruction as an open challenge.


\subsection{Vector Graphics Generation}
Vector graphics generation studies how to synthesize scalable vector representations from raster inputs, typically in the form of SVG programs composed of geometric primitives and drawing commands. Classical vectorization tools, such as Potrace~\cite{selinger2003potrace}, recover smooth curves from raster bitmaps through shape-driven tracing. Neural and differentiable methods recast this problem as learning vector representations: DiffVG~\cite{li2020differentiable} introduces a differentiable rasterizer enabling gradient-based editing and learning over vector primitives, while Neural methods recast vectorization as learning problems: DiffVG~\cite{li2020differentiable} enables gradient-based editing via differentiable rendering; DeepSVG~\cite{carlier2020deepsvg} disentangles shapes from drawing commands for icon generation; Im2Vec~\cite{reddy2021im2vec} and LIVE~\cite{ma2022towards} synthesize vectors from raster supervision using differentiable rendering and progressive layer-wise fitting respectively. While these methods improve fidelity and flexibility over classical tracing, they largely treat vector graphics as collections of geometric primitives, with limited modeling of semantic object structures.

More recently, multimodal foundation models have reframed SVG generation as a code-generation task. StarVector~\cite{rodriguez2025starvector} employs a multimodal large language model to generate SVG code from images and text, supporting primitives such as ellipses, polygons, and text, and introduces SVG-Stack and SVG-Bench for broader evaluation. LLM4SVG~\cite{xing2025empowering} leverages learnable semantic tokens, a modular architecture for vector instructions, and large-scale SVGX datasets to reduce hallucinations and improve semantic alignment. RLRF~\cite{rodriguez2025rendering} further enhances autoregressive SVG generation by optimizing rendered outputs with reinforcement learning, combining rewards for reconstruction fidelity, semantic consistency, and code efficiency.

Despite these advances, prior work targets generic graphics such as icons and illustrations, without handling chart-specific structures such as axes, legends, and grouped marks.

\subsection{Visualization Reuse and Modification}
A lot of work focuses on the reuse and augmentation of existing visualizations, as designers often find it more natural to modify existing graphics than to start from scratch~\cite{bigelow2016iterating, bako2022understanding}. While traditional template-based systems offer a user-friendly entry point, they are frequently limited by the fixed expressivity and quantity of templates provided by developers. To overcome these constraints, research has investigated how to transform existing visualizations into reusable templates without manual intervention. 
D3 Deconstructor~\cite{harper2017converting} and iVoLVER~\cite{mendez2016ivolver} pioneered data-driven reuse from D3-based and heterogeneous sources;
Ivy~\cite{mcnutt2021integrated} and Cui et al.~\cite{cui2021mixed} further converted specifications into parameterized templates and supported mixed-initiative infographic retargeting.

Building on these foundations, DIVI~\cite{snyder2023divi} and Mystique~\cite{chen2023mystique} deconstruct SVG charts to recover semantic components such as axes, legends, and data encodings. Mystique further decomposes complex rectangle-based layouts into mark groups, spatial relationships, data encodings, and graphical constraints. By formalizing these relationships without source data or D3.js metadata, it enables the reuse of advanced layouts like small multiples and nested groupings. Structural recovery also supports accessibility; Choi et al.~\cite{choi2019visualizing} showed that understanding a chart’s structure is essential for tactile graphics. Recent datasets like VisAnatomy~\cite{chen2025visanatomy} provide fine-grained labels for element roles and hierarchical grouping. DataWink~\cite{xie2025datawink} allows users to adapt high-quality SVG visualizations using LLMs to extract data encodings and edit appearance via a conversational interface and on-demand widgets. 

Despite these advances, the vast majority of reuse systems assume access to structured vector sources and cannot operate on raster images, a gap that Chart2SVG directly addresses.


%% file: sections/3_method.tex
\section{\ourmodel}
We study the problem of reconstructing editable SVG charts from raster chart images. 
Our approach, \ourmodel, combines a chart-oriented data preparation pipeline with a vision-language reconstruction model trained in two stages. We first build a large-scale paired corpus of raster charts and SVG code, then canonicalize heterogeneous raw SVGs into a unified representation. This format is more regular for learning while preserving the structural cues needed for downstream editing. On top of this representation, we adapt a pretrained vision-language model with SVG-specific semantic tokens and train it using supervised fine-tuning (SFT) followed by rendering-aware reinforcement learning (RL).

\subsection{Data Preparation Pipeline}
High-quality SVG supervision is essential for chart-to-SVG reconstruction. 
However, raw web SVGs are noisy, tool-dependent, and often poorly aligned with the rendered chart appearance. 
We therefore construct a paired chart corpus and transform all SVG targets into a canonical chart-oriented representation before model training.

\paragraph{Data Source.}
Our primary requirement is real-world chart data featuring aligned raster images and SVG code. However, existing datasets are limited in scale: VisAnatomy~\cite{chen2025visanatomy} contains only 942 real-world pairs curated for semantic diversity, while the REV~\cite{poco2017reverse} corpora include fewer than 500 charts from Quartz/Atlas and academic papers. To address this scarcity, we construct \textit{Beagle+}, our main training corpus, based on Beagle~\cite{battle2018beagle}. This collection comprises over 41K SVG visualizations mined from five web-based repositories across 24 chart types. We further broaden coverage by crawling 276 additional examples from the ECharts~\cite{li2018echarts} gallery. Notably, we exclude the D3 subset from Beagle; its long tail of bespoke designs and irregular structural patterns makes it less suitable for establishing a stable reconstruction model during the initial training stage.

While SVG serves as a versatile vector graphics format, it is semantically agnostic and possesses no inherent concept of chart elements such as bars or axes. Consequently, visually similar charts may be encoded with vastly different document object model (DOM) structures, primitive choices, style systems, and transform hierarchies depending on the authoring tool used. In our corpus, some tools generate deeply nested <g> structures while others produce much flatter SVG trees; similarly, some encode appearance through inline attributes, whereas others rely heavily on global <style> blocks and CSS selectors. We also observe substantial variation in geometric syntax, including the use of absolute versus relative path commands and multi-level transform chains.

Such heterogeneity introduces significant noise for sequence modeling, as the model must learn tool-specific implementation details alongside chart semantics. However, simply flattening SVGs into geometry-only representations using tools like SVGO~\cite{svgo} is undesirable, as it discards the grouping and primitive information essential for downstream editing. We therefore seek a representation that suppresses accidental cross-tool variation while preserving the structural cues that maintain chart editability.

\paragraph{SVG Charts Canonicalization.}
To reduce this heterogeneity without sacrificing editability, we canonicalize all raw SVG charts into a unified chart-oriented representation.
Our goal is not generic SVG compression, but a representation that is both more regular for sequence modeling and still structurally meaningful for downstream chart editing.
To this end, we preserve chart-relevant structure whenever possible, including meaningful \texttt{<g>} hierarchies, primitive tags, and useful \texttt{class} attributes, rather than collapsing the entire chart into anonymous paths.

\begin{enumerate}
    \item \textit{Canvas Standardization:} injecting standard namespaces and a \texttt{viewBox}, and unifying the canvas dimensions to ensure consistent rendering across diverse environments;
    \item \textit{Style Normalization:} converting tool-specific CSS styles into element-level attributes and removing inline style blocks to eliminate priority conflicts;
    \item \textit{Geometric Canonicalization:} standardizing path representations using relative coordinates and flattening nested \texttt{transform} operations to simplify geometric structures, thereby facilitating position computation and local editing; and 
    \item \textit{Structural Distillation:} removing redundant placeholder elements, unnecessary attributes, and embedded bitmap content, while preserving the original hierarchical structure and primitive semantics (e.g., \ \texttt{line},  \texttt{rect},  \texttt{circle}) to improve clarity without sacrificing meaning.
\end{enumerate}

The steps of style normalization and geometric canonicalization enable consistent and efficient global modifications of the converted SVG files, while the structural distillation step ensures that the final representation remains lightweight and semantically transparent for downstream sequence modeling. Compared with raw web-generated SVGs, the resulting representation is substantially more compact, typically reducing the size to 46\%--72\% of the original. 

Since real-world datasets often contain overly dense and large SVG files (e.g., maps or large-scale scatter plots), it is challenging for the model to learn complete and well-formed SVG sequences. To address this, we filter out samples whose token length exceeds the maximum context length (8192), resulting in a curated dataset of 33K valid examples, which we denote as \textit{Beagle+}. 
\new{Beagle+ is derived primarily from the Beagle dataset~\cite{battle2018beagle} and is produced through our chart-oriented canonicalization pipeline.}

\subsection{Network Architecture and Training}

\begin{figure*}[t]
  \centering
  \includegraphics[width=1.0\textwidth]{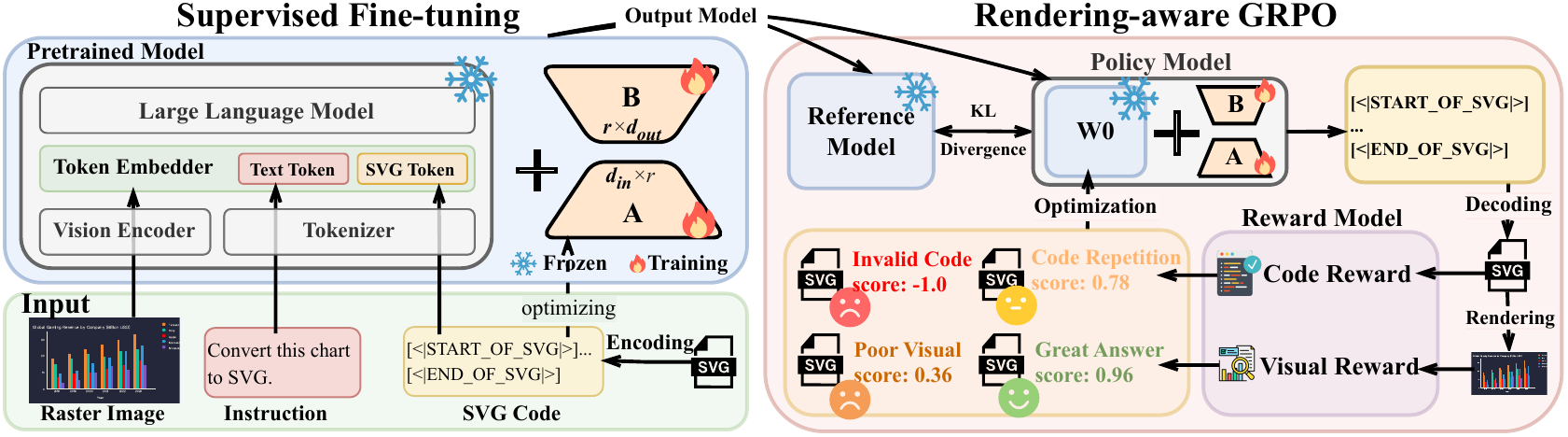}\vspace{-2mm}
  \caption{Training pipeline. (Left) SFT: a vision encoder and LLM are aligned via parameter-efficient adapters to generate structured SVG sequences. (Right) Rendering-aware GRPO: SVG candidates are evaluated by code validity and visual fidelity rewards, with KL divergence against a reference model for regularization.}\vspace{-4mm}
  \label{fig:training}
\end{figure*}

As shown in Fig.~\ref{fig:training}, our reconstruction model combines a pretrained vision-language backbone, a structured SVG tokenization scheme, and a two-stage training strategy. 
The first stage learns chart-to-SVG generation by supervised fine-tuning (SFT), while the second stage refines the model using rendering-aware reinforcement learning to better align generated SVGs with their rendered appearance.

\paragraph{Problem Formulation and Backbone.}
We formulate chart-to-SVG reconstruction as a conditional sequence generation task. 
Given a raster chart image $I$, the model predicts a canonicalized SVG token sequence $\hat{S} = [s_1, \dots, s_L]$ in an autoregressive manner. 
Concretely, the visual encoder extracts image features from $I$, which are projected into the language model as visual tokens. The model then decodes SVG tokens sequentially conditioned on both the visual context and previously generated tokens:
\begin{equation}
    P(\hat{S} \mid I) = \prod_t P(s_t \mid I, s_{<t}).
\end{equation}
This formulation enables unified modeling of chart structure, geometry, and styling within a single VLM framework.

We instantiate our model on top of Qwen3-VL~\cite{bai2025qwen3}, a large-scale vision-language model designed for unified multimodal understanding. Qwen3-VL is well-suited to this task, as chart reconstruction requires capturing both fine-grained visual details and long-range structural relationships. In this context, the model must accurately interpret subtle elements such as ticks and labels while preserving the global semantic coherence necessary to produce a functional and editable SVG.


\paragraph{SFT with SVG Grammar Tokens.}
To adapt the backbone to chart reconstruction, we extend the tokenizer with 115 SVG-specific special tokens and perform supervised fine-tuning on paired raster chart-SVG data.
Rather than treating SVG as plain text, we represent it with structured token units designed to better match the semantic organization of chart graphics.
Formally, we extend the tokenizer vocabulary as
\begin{equation} 
\mathcal{V}
=
\mathcal{V}_{\text{base}}
\cup
\mathcal{V}_{\text{struct}}
\cup
\mathcal{V}_{\text{prim}}
\cup
\mathcal{V}_{\text{attr}}
\cup
\mathcal{V}_{\text{geom}},
\end{equation}
where \(\mathcal{V}_{\text{base}}\) denotes the original tokenizer vocabulary of the pretrained Qwen3-VL model, and the added SVG-specific tokens are organized into four groups, i.e., structural containers, graphical primitives, visual channel attributes, and geometric opcodes, following the chart-oriented taxonomy of VisAnatomy~\cite{chen2025visanatomy}, excluding element-type semantics and reference elements:
\begin{enumerate}
    \item \emph{Structural containers and anchors} ($\mathcal{V}_{\text{struct}}$), including tokens such as \texttt{<svg>}, \texttt{<g>}, \texttt{<defs>}, and \texttt{<clipPath>}, which capture document scope and grouping structure;
    \item \emph{Graphical primitives} ($\mathcal{V}_{\text{prim}}$), including \texttt{<rect>}, \texttt{<circle>}, \texttt{<line>}, \texttt{<path>}, \texttt{<text>}, and \texttt{<use>}, which represent the basic chart elements;
    \item \emph{Visual channel attributes} ($\mathcal{V}_{\text{attr}}$), including geometric and stylistic tokens such as \texttt{x}, \texttt{y}, \texttt{width}, \texttt{height}, \texttt{fill}, \texttt{stroke}, \texttt{font-size}, \texttt{opacity}, and \texttt{transform}; and
    \item \emph{Geometric opcodes} ($\mathcal{V}_{\text{geom}}$), including path commands such as \texttt{M}, \texttt{L}, \texttt{C}, \texttt{Q}, and \texttt{Z}, which encode complex trajectories such as polylines, curves, and area boundaries.
\end{enumerate}
Under this tokenization scheme, the model predicts a canonicalized SVG sequence composed of semantically meaningful units rather than fragmented XML strings.
A complete definition of all SVG-specific special tokens is provided in the supplementary materials.

For each newly introduced token, we assign its embedding as the average embedding of a short natural-language description of its SVG meaning, so that semantically related tokens occupy nearby positions in the pretrained language space.
Concretely, if a token is associated with a description sequence \(\{w_j\}_{j=1}^{n}\), its embedding is defined as
\begin{equation}
\mathbf{e}
=
\frac{1}{n}\sum_{j=1}^{n}\mathbf{E}(w_j),
\end{equation}
where \(\mathbf{E}(\cdot)\) denotes the pretrained embedding lookup of the base tokenizer.
We keep the embedding layer fixed during training.

We fine-tune the model with the standard autoregressive objective to maximize the likelihood of the target SVG token sequence given the input chart image.
Instead of fully tuning the entire backbone, we use LoRA~\cite{hu2022lora} to update a small set of low-rank adaptation parameters:
\begin{equation}
W = W_0 + \Delta W, \qquad \Delta W = BA,
\end{equation}
where $W_0$ is the frozen pretrained weight and $B,A$ are trainable low-rank factors.
This design substantially reduces training cost and memory usage, while preserving the strong multimodal priors of the pretrained Qwen3-VL model.
The SFT stage teaches the model both the grammar of SVG and the mapping from chart appearance to structured vector representations, including hierarchical grouping, graphical primitives, and encoding attributes.

\paragraph{Rendering-aware GRPO.}
While SFT optimizes token-level likelihood, it does not directly enforce the rendered quality of the generated SVG.
In practice, small token-level errors, such as misplaced coordinates or duplicated primitives, can lead to large perceptual artifacts after rendering.
To better align training with the actual reconstruction objective, we further refine the model using rendering-aware reinforcement learning based on Group Relative Policy Optimization (GRPO)~\cite{shao2024deepseekmath}.
Given an input chart image $I$, the current policy $\pi_{\theta}$ samples a group of $G$ candidate SVG sequences $\{S_i\}_{i=1}^{G}$.
Each candidate is rendered into a raster image \(\tilde{I}_i\), and rewards are computed from both the SVG code and the rendered output.

We first define a \textbf{code-level reward} to encourage syntactically valid and compact SVG generation. Unlike natural language, SVG must satisfy strict structural constraints to be executable.
We therefore assign a validity reward $R_{\text{valid}}(S_i)$ based on whether the generated SVG can be successfully parsed and rendered.
In addition, we penalize unnecessarily long or redundant sequences to discourage degenerate behaviors such as repeatedly drawing overlapping shapes. The code-level reward is defined as
\begin{equation}
R_{\text{code}}(S_i) = R_{\text{valid}}(S_i) - \lambda_{\text{len}}\,\mathcal{P}_{\text{len}}(S_i).
\end{equation}

We then define a \textbf{visual reward} that directly measures how well the rendered SVG matches the target chart.
Since charts are structured and information-dense, no single image metric is sufficient. We therefore combine three complementary measures:
(i) \emph{Foreground Intersection over Union} (IoU), which evaluates overlap on non-background regions and focuses on chart content such as marks, axes, and text;
(ii) \emph{Structural Similarity Index} (SSIM), which captures local structural consistency and is sensitive to thin lines and textual elements; and 
(iii) \emph{Peak Signal-to-Noise Ratio} (PSNR), which measures overall pixel-level fidelity and color consistency. The visual reward is
\begin{equation}
R_{\text{vis}}(\tilde{I}_i, I)=
\frac{1}{3}\Big(
\mathrm{IoU}_{\mathrm{fg}}(\tilde{I}_i, I)
+
\mathrm{SSIM}(\tilde{I}_i, I)
+
\mathrm{PSNR}(\tilde{I}_i, I)
\Big).
\end{equation}

\begin{figure*}[t]
  \centering
  \includegraphics[width=1.0\textwidth]{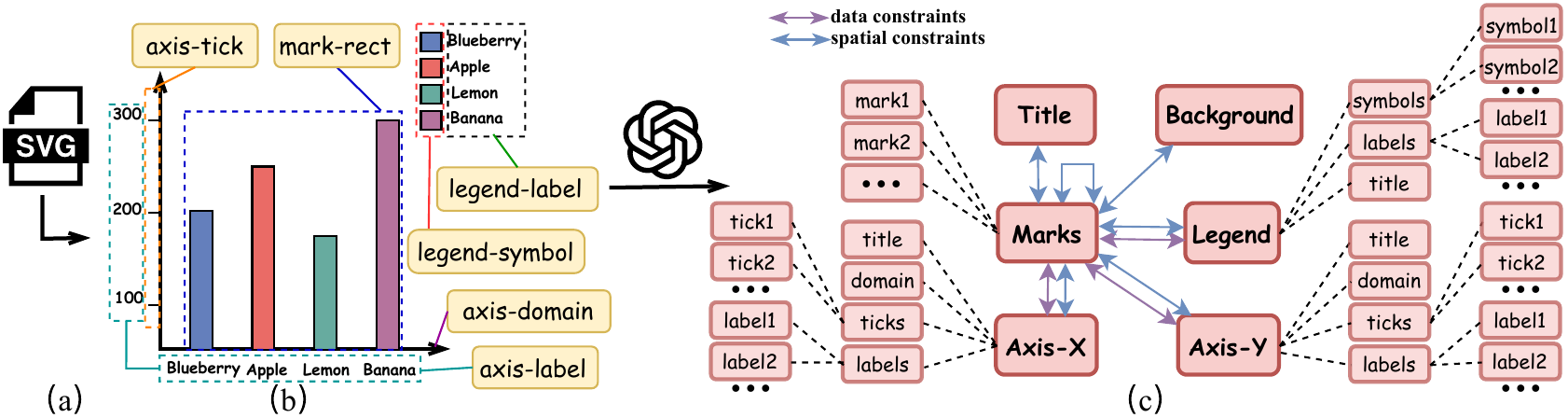}\vspace{-3mm}
\caption{CSG construction pipeline. From a reconstructed SVG (a), chart components are extracted (b) and organized into a graph (c) with data constraints (governing mark-axis-legend relationships) and spatial constraints (defining layout relative to marks).}
  \label{fig:scene-graph}\vspace{-6mm}
\end{figure*}

Note that our visual reward design differs from RLRF~\cite{rodriguez2025rendering} because chart reconstruction places greater emphasis on precise structural fidelity than on high-level semantic similarity. 
In generic SVG generation, perceptual rewards are useful for encouraging semantic alignment, but for charts, small geometric errors in marks, ticks, labels, or legends can alter the conveyed data while remaining semantically similar at a coarse level. 
We therefore use chart-oriented visual rewards based on foreground IoU, SSIM, and PSNR, so that optimization better reflects the fine-grained geometric requirements of editable chart SVGs. \new{The detailed computation of the reward functions is provided in Appendix~\ref{grpo}.}

The final reward balances code-level validity and visual fidelity by combining the two components:
\begin{equation}
R(S_i, I) = \lambda_{\text{code}} R_{\text{code}}(S_i) + \lambda_{\text{vis}} R_{\text{vis}}(\tilde{I}_i, I).
\end{equation}
Here, $\lambda_{\text{code}}$ and $\lambda_{\text{vis}}$ control the relative importance of code correctness and visual accuracy.
The code-level reward ensures that outputs remain executable, compact, and structurally well formed, while the visual reward encourages faithful reconstruction of chart geometry and appearance. 
In practice, we assign greater weight to the visual term, while retaining a non-negligible code reward to avoid invalid or excessively verbose SVG outputs.

Following GRPO~\cite{shao2024deepseekmath}, we sample a group of $G$ candidate SVG sequences for each input and normalize their rewards within the group to obtain relative advantages $A_i$. Specifically, $A_i$ is computed by standardizing rewards across the group, encouraging the model to prefer relatively better candidates rather than relying on absolute reward scales. 
The policy is then updated using the clipped objective
\begin{equation}
\mathcal{L}_{\mathrm{GRPO}}
=
-\frac{1}{G}\sum_{i=1}^{G}
\min\!\left(
r_i A_i,\;
\mathrm{clip}(r_i, 1-\delta, 1+\delta)A_i
\right),
\end{equation}
where $r_i=\pi_{\theta}(S_i\mid I)/\pi_{\theta_{\mathrm{old}}}(S_i\mid I)$ is the likelihood ratio between the current and previous policies, and $\delta$ controls the clipping range to stabilize updates.

By optimizing rewards defined on rendered outputs rather than only token sequences, this stage better aligns training with the perceptual and structural objectives of chart reconstruction, leading to higher visual fidelity and more reliable SVG structure.

%% file: sections/4_application.tex
\section{Semantic Structure-Aware Chart Manipulation}
\label{sec:semantic_abs}
The goal of Chart2SVG is not only to reconstruct charts as SVGs, but also to make vectorized charts directly editable through language-guided manipulation. This interaction paradigm empowers users to issue high-level instructions rather than manually modifying low-level vector code. Prior work such as DirectGPT~\cite{masson2024directgpt} improves the efficiency of vector graphic editing by combining natural language with direct manipulation. However, it operates on vector primitives, where users select elements and apply isolated edits. While effective for generic vector graphics, this approach is insufficient for charts, which are inherently governed by strict structural regularity and semantic organization.

The primary challenge in chart understanding lies in bridging the semantic gap between low-level SVG primitives and high-level chart constructs. This challenge is twofold. First, individual primitives are fundamentally ambiguous in isolation. A single \texttt{<line>} might denote an axis domain, a gridline, or a tick mark, just as a \texttt{<rect>} could represent a data bar, a legend entry, or a background bounding box. 
Recent systems such as DIVI~\cite{snyder2023divi} and Mystique~\cite{chen2023mystique} provide robust heuristics to map raw SVG elements to high-level chart components, effectively addressing this labeling problem. 

Second, chart elements are not independent; they are tightly coupled through complex structural and semantic dependencies. Modifying one component invariably necessitates coordinated updates to others. For instance, altering a data series requires updating the legend, and adjusting layout properties like bar width dictates shifts in spacing and alignment across the entire chart. Capturing these interdependencies remains largely unaddressed.

To bridge this gap, we build on the heuristically labeled components from DIVI and leverage large language models (LLMs) to automatically infer additional structural and semantic rules. Based on these relationships, we transform the labeled SVG into an explicit \emph{Chart Structure Graph} (CSG), formally defined as $G = (V, E)$, as illustrated in Fig.~\ref{fig:scene-graph}. Serving as a semantic interface between the SVG and the model, the CSG allows users to interact directly with structured chart components, their mapped attributes, and topological dependencies rather than disjoint low-level primitives. 

\subsection{Constructing Chart Structure Graphs}
\label{subsec:CSG}
\new{Before constructing the CSG, we first annotate the reconstructed SVG with chart-level semantic roles using DIVI's scene-graph vocabulary~\cite{snyder2023divi} as the base component schema and its heuristic grouping rules. Specifically, low-level SVG elements are assigned role-instance identifiers, such as marks, axis ticks, legend symbols, and text labels, which serve as semantic anchors for grouping elements and locating edit targets.}
\new{Given the tagged SVG and the schema, an LLM then constructs the vertex set V by grouping tagged elements that share the same semantic role or belong to the same component group and instantiating them as concrete chart-component nodes.}
\new{Each node is assigned a typed component label and a semantic identifier, which contains fields such as \texttt{data\_key}, \texttt{scale\_ref}, \texttt{svg\_selector}. Because the DIVI schema restricts the node vocabulary to a fixed set of chart component types, the LLM is prevented from introducing unsupported role types.}
\new{In parallel, each node retains the corresponding SVG element set or subtree as its low-level realization. This separation allows the CSG to reason over chart semantics at the graph level while preserving a concrete link to the underlying SVG elements, so that graph-level operations can be translated back into concrete SVG attribute updates.}



\new{The next step is to define the edge set $E$, which captures semantic and geometric dependencies among chart-component vertices. Given the instantiated vertices, their semantic attributes, and their associated SVG element sets or subtrees, we prompt an LLM to instantiate edges under a predefined constraint schema rather than freely generating arbitrary relationships. Each edge connects one or more source vertices to a target vertex and is assigned a constraint type, the affected visual attributes, and the semantic basis for the dependency. We consider two categories of constraints: data constraints and spatial constraints.}

\vspace{0.5em}
\noindent\textbf{Data Constraints.}
\new{These directed edges encode how data semantics determine visual encodings. They are instantiated when vertices share data keys, scale references, legend-category bindings, or value-dependent geometric attributes.}
The LLM infers from three core rules:
\begin{itemize}[leftmargin=*]\itemsep0pt
    \item \emph{Axis Scaling:} Constrains the mapping between the data domain and the coordinate system. Changes in the data range require rescaling the axes and remapping all dependent elements.
    \item \emph{Value Proportionality:} Ensures that geometric attributes (e.g., height, length, or area) remain proportional to the underlying data values.
    \item \emph{Category Consistency:} Maintains uniform visual encoding (e.g., color, texture, shape) for all elements within the same data category.
\end{itemize}

\vspace{0.5em}
\noindent\textbf{Spatial Constraints.}
\new{These edges encode layout dependencies among vertices based on their bounding boxes, relative positions, and grouping relationships.}
The LLM reasons based on four key rules:
\begin{itemize}[leftmargin=*]\itemsep0pt
    \item \emph{Element Alignment:} Ensures graphical elements and gridlines align precisely with axis ticks.
    \item \emph{Group Spacing:} Maintains uniform spacing among related elements, requiring coordinated updates within groups.
    \item \emph{Stacking Order:} Encodes cumulative positioning, where each element depends on preceding elements in the stack.
    \item \emph{Overlap Prevention:} Avoids overlaps between text and graphical elements by triggering positional adjustments when conflicts arise.
\end{itemize}

For example, in Fig.~\ref{fig:scene-graph}, there is a data constraint between marks and axes: their coordinates are jointly determined by the same scale, so their positions remain consistent. Similarly, a data constraint exists between marks and the legend: visual style (primarily color) must be consistent, ensuring that elements of the same data category share the same appearance. Regarding spatial constraints, marks maintain layout relationships with each other and are typically aligned with axes (e.g., the bottoms of bars align with the x-axis in a bar chart). The title is positioned relative to the marks.

The chart structure graph enables consistent chart manipulation. When a user issues a high-level editing command, the system traverses the graph to propagate updates along these dependencies, preserving both semantic correctness and visual coherence. To maintain consistency under layout changes, the CSG employs a bidirectional constraint propagation mechanism: local modifications that increase spatial demand (e.g., enlarging elements or adding data) propagate upward, expanding axes and the canvas to prevent overlap or clipping, while global constraints (e.g., reducing canvas size) propagate downward, compressing element sizes, spacing, and typography. This bidirectional propagation ensures both local edits and global adjustments yield structurally valid and visually consistent chart layouts.
\new{The CSG inference prompt template is provided in Appendix~\ref{sec:CSG_prompt}.}

\subsection{CSG-aware Chart Manipulation}
\label{subsec:CSG-editing}
CSG serves as the core intermediate representation for chart manipulation. 
\new{Given a natural-language editing instruction, the LLM first grounds the instruction to one or more target vertices in the CSG using their component types, semantic attributes, textual labels, and geometric properties. This graph-level grounding restricts edits to existing chart components rather than arbitrary SVG primitives. The system then follows the dependency edges connected to the target vertices to identify other components that should be updated jointly, such as axes, legends, or mark groups. Based on this affected subgraph, the LLM produces a graph-level edit plan that specifies the vertices to update and the corresponding attribute changes. The plan is then executed on the SVG element sets or subtrees associated with these vertices, and the modified SVG DOM is written back as editable SVG code.}


For example, the command of ``increase the bar width for the banana data in Fig.~\ref{fig:scene-graph}(b)'' triggers a graph operation that updates the relevant vertices representing the bars. Thanks to bidirectional constraint resolution, this local modification automatically propagates spatial adjustments to dependent vertices, such as expanding axes, shifting intra-group spacing, and updating legends. Similarly, ``highlight all points in Category A'' synchronizes attribute updates across all vertices in that data group, respecting the \emph{category binding} rules encoded in the CSG.

\new{In contrast to naive SVG editing, where a model must infer element coupling from raw SVG structure and geometry, CSG-guided editing exposes chart-component dependencies explicitly, allowing the LLM to plan coordinated updates over semantically related components. Our implementation uses an off-the-shelf LLM executor (GPT-5.1) for graph-edit planning, but the CSG interface is model-agnostic and can be paired with other models.}

%% file: sections/5_evaluation.tex
\input{tables/recon_id}

\input{tables/recon_ood}

\section{Evaluation}
\label{sec:eval}

\subsection{Metrics and Baselines}
\label{metrics}

\paragraph{Metrics.}
We evaluate reconstruction quality by rasterizing both the predicted SVG and the ground-truth SVG at the same resolution and comparing the resulting images.
Since chart images often contain large uniform background regions, evaluating similarity over the entire canvas can be misleading: a prediction may match the background well while still failing to reconstruct the chart itself.
We therefore adopt a foreground-aware evaluation protocol tailored to chart images.

Specifically, we estimate the background color from the top-left pixel of the rendered image and identify foreground pixels as those whose color differs from the background by more than a small threshold (\texttt{tol}=5).
The resulting foreground mask is used to automatically crop each chart to its content region before computing image similarity metrics.
This reduces the influence of large blank margins and makes the evaluation more sensitive to the reconstructed chart elements.

Let $I$ and $\hat{I}$ denote the cropped rasterized ground-truth and predicted images, respectively, and let $M$ and $\hat{M}$ denote the corresponding foreground masks.
We report the following metrics:
\begin{enumerate}
    \item \textbf{PSNR}, computed from the mean squared pixel error between $I$ and $\hat{I}$, which measures low-level reconstruction fidelity. Specifically, it is computed as $\mathrm{PSNR} = 10 \log_{10}(\mathrm{MAX}^2 / \mathrm{MSE})$, where $\mathrm{MAX}=255$ for 8-bit raster images.

    \item \textbf{SSIM}, which measures structural similarity between the reconstructed chart and the ground truth by comparing local luminance, contrast, and structural patterns.

    \item \textbf{LPIPS}, computed with the AlexNet backbone, which measures perceptual dissimilarity between the reconstructed chart and the ground truth in a deep feature space.

\item \textbf{Foreground IoU}, defined as $\mathrm{IoU}_{\mathrm{fg}} = |M \cap \hat{M}| / (|M \cup \hat{M}| + \epsilon)$, measures overlap between predicted and ground-truth chart foregrounds while reducing background influence.

\item \textbf{Edge Consistency}, computed from Canny edge maps with light dilation and smoothing, quantifies agreement between chart edges, emphasizing thin but semantically critical elements like axes, ticks, gridlines, and line marks.  

\item \textbf{Failure Rate}, the fraction of predictions that cannot be rendered or whose background occupies more than $97\%$ of the image, indicating failure to produce meaningful chart.
\end{enumerate}

The first three metrics (PSNR, SSIM, and LPIPS) are standard in prior SVG generation models like StarVector~\cite{rodriguez2025starvector}, while the latter three are particularly useful for chart-specific evaluation. For PSNR, SSIM, and Foreground IoU, higher values indicate better performance, whereas lower LPIPS and Failure Rate are better.

\paragraph{Baselines.}
We compare \ourmodel against four representative baselines.
OmniSVG~\cite{yangomnisvg} is an end-to-end SVG generation model using discrete SVG tokenization for multimodal synthesis.
StarVector~\cite{rodriguez2025starvector} is a multimodal LLM that directly produces compilable SVG code from images or text.
ChartCoder~\cite{zhao2025chartcoder} is a dedicated chart-to-code model using a code-oriented backbone for executable chart restoration.
We further include GPT-4o as a strong proprietary baseline capable of accepting image inputs and generating structured code outputs.

\paragraph{Implementation Details.}
We train two variants of \ourmodel based on Qwen3-VL~\cite{bai2025qwen3}, with 4B and 8B parameters, sharing the same architecture and training pipeline.
Input images are resized to $512 \times 512$, with maximum output length set to 8,192 tokens after canonicalization.
For SFT, we use AdamW with learning rate $1\times10^{-4}$, 2 epochs, batch size 16, and LoRA rank $r=8$.
For RL, we use group size $G=4$, clipping parameter $\delta=0.05$, sampling temperature $T=1.0$, and $\lambda_{\text{code}}=0.2$, $\lambda_{\text{vis}}=0.6$, $\lambda_{\text{len}}=0.2$.
All models are trained on 8 NVIDIA L20 GPUs for approximately 9 hours.
\new{More detailed baseline configuration  and comparative analysis are provided in the Appendix.}

\subsection{Quantitative Results}
\label{sec:quantitative_results}

\paragraph{In-Distribution Test.}
Table~\ref{tab:indistribution} reports in-distribution results on three held-out subsets in Beagle+, generated using ChartBlocks, Fusion, and Plotly.
Overall, both \ourmodel variants substantially outperform prior baselines across the three subsets, and, most importantly, achieve a zero failure rate on all of them.
This indicates that the proposed chart-oriented representation and training strategy improve not only visual similarity, but also the basic validity of generated SVG charts.


On ChartBlocks, Chart2SVG-8B achieves the best result on all metrics (15.56 PSNR, 0.71 SSIM, 0.26 LPIPS) with zero failure rate. On Fusion and Plotly, where StarVector leads on PSNR due to its sensitivity to 
spatial alignment, Chart2SVG models remain strongest on structural metrics and are the only methods with zero failures across all three subsets.
For chart images, where many semantically important elements such as axes, ticks, and line marks are thin structures, even minor positional offsets can lead to substantial pixel error. 
As a result, a method may obtain high PSNR by matching coarse raster appearance, yet still reconstruct chart structure less faithfully than our method, as also reflected in the qualitative comparisons.

The \emph{Plotly} subset is the most challenging.
Here, StarVector achieves the highest PSNR, but both \ourmodel models remain strongest on most other metrics and are again the only methods with zero failures.
\ourmodel-4B performs best on SSIM and $\mathrm{IoU}_{\mathrm{fg}}$, while \ourmodel-8B performs best on LPIPS and edge consistency.
This split suggests that the two model scales make slightly different trade-offs on Plotly charts, but both remain substantially more reliable than prior baselines.

\paragraph{Out-of-Distribution Test.}
Table~\ref{tab:ood} reports out-of-distribution results on the VisAnatomy~\cite{chen2025visanatomy} test set.
All neural methods experience a noticeable performance drop compared with the in-distribution setting, confirming that cross-dataset generalization remains challenging. 
Adobe Illustrator achieves high visual fidelity metrics owing to its pixel-faithful path tracing, but produces no semantic structure and a zero failure rate only because it always renders something, 
regardless of chart correctness.
Among learning-based methods, \ourmodel-8B achieves the best results across all five metrics and maintains a failure rate of only 0.78\%, outperforming GPT-4o by 12.28\% in PSNR and 24.48\% in edge consistency, and remaining competitive with the stronger proprietary 
models o3 and Gemini-2.5-Flash while producing semantically structured, directly editable SVG output.

\input{tables/recon_data}
\new{
\paragraph{Data Value Accuracy.}
Table~\ref{tab:numerical_by_type_ood} reports data accuracy on VisAnatomy by chart type, measuring array-length matching rate (Len.) and numerical accuracy over successfully matched data (Acc.). StarVector and OmniSVG produce almost no valid output across all chart types. Among learning-based methods, Chart2SVG-8B achieves superior accuracy: on bar charts it leads all baselines in Acc.\ (0.1807 vs.\ at most 0.1545), on pie charts it is competitive with o3 and GPT-4o while Gemini-2.5-Flash achieves the highest Acc.\ (0.1664),
and on line charts Chart2SVG-8B achieves the highest Acc.\ (0.2173) among all models.}
\new{These results suggest that Chart2SVG better preserves quantitative structure when the recovered data arrays can be aligned, although array-length mismatches remain a major source of error.}

\paragraph{\new{Summary.}}
\new{Taken together, \ourmodel consistently improves reconstruction fidelity and robustness across both in-distribution and out-of-distribution settings, achieving the lowest failure rate among all learning-based methods and remaining competitive with stronger proprietary models. Note that although Adobe Illustrator achieves strong reconstruction performance, its output SVG code consists primarily of fitted Bézier-curve coordinates, which lack explicit chart semantics and are therefore difficult to interpret, edit, or reuse for downstream chart understanding and data extraction.}

\subsection{Qualitative Analysis}

\begin{figure*}[!t]
  \centering
  \includegraphics[width=0.94\textwidth]{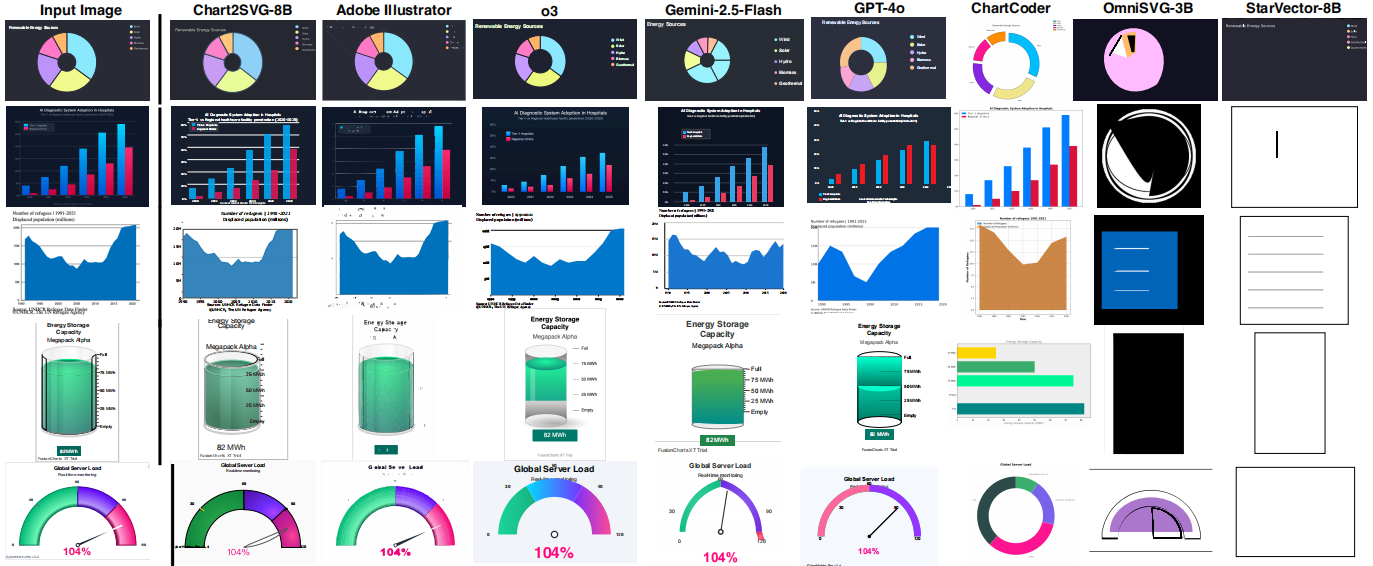}\vspace{-2mm}
\caption{Qualitative comparison of our model with general-purpose and state-of-the-art chart and SVG foundation MLLMs on chart vectorization.}\vspace{-3mm}
  \label{fig:qualitative}
\end{figure*}

We present a qualitative comparison in Fig.~\ref{fig:qualitative} spanning standard charts and highly customized pictorial charts across three dimensions: \emph{structural fidelity}, \emph{data accuracy}, and \emph{aesthetic preservation}.

\paragraph{Structural Fidelity.}
Accurate reconstruction requires both precise geometry and correct spatial organization. GPT-4o often struggles with continuous geometry, producing jagged curves in doughnut charts and unintended spacing between grouped bars. ChartCoder preserves the basic structure of standard charts but struggles to generalize to complex or non-canonical layouts. StarVector-8B and OmniSVG-3B frequently yield fragmented or incomplete outputs, particularly for charts with long token sequences or multi-element compositions. \new{Adobe Illustrator achieves high geometric fidelity via path tracing, but all text labels and titles are rendered illegible or entirely missing, as anti-aliased character outlines collapse into noise rather than recoverable typography.} In contrast, \ourmodel consistently generates well-aligned primitives and preserves strict layout coherence, capturing both local geometric details and global compositional structures, even for highly customized pictorial charts where spatial precision is essential for readability.

\paragraph{Aesthetic Preservation.}
ChartCoder reverts to generic library defaults, with severe degradation on pictorial charts. GPT-4o struggles with complex shading and text alignment. \ourmodel better preserves custom color palettes, advanced effects, and text layout, maintaining visual fidelity across both standard and metaphorical mark styles.

\paragraph{Summary.}
\ourmodel achieves a superior balance across structure, data, and aesthetics, producing robust reconstructions that bridge the gap between raster images and \new{semantically structured vector representations that support programmatic editing workflows.}

\subsection{Ablation Studies}

\input{tables/ablate_new}

Table~\ref{tab:ablation} reports ablation results on the 4B model, where the full \ourmodel-4B serves as the reference, and one component is removed at a time. Overall, both chart-specific SVG tokenization and the rendering-aware GRPO objective substantially contribute to reconstruction quality and output validity.

\paragraph{Effect of SVG Grammar Tokens.}
Removing SVG grammar tokens degrades all metrics, with PSNR dropping from 14.02 to 12.23 and failure rate rising from 0.52\% to 14.81\%, confirming that chart-oriented tokenization is critical for visual fidelity and rendering reliability.

\paragraph{Effect of GRPO Training.}
Omitting GRPO causes the largest drop overall: PSNR falls from 14.02 to 6.81 and SSIM from 0.6951 to 0.4672. Notably, $\mathrm{IoU}_{\mathrm{fg}}$ slightly increases without GRPO, highlighting that foreground overlap alone does not capture reconstruction quality---a prediction can cover the foreground yet produce incorrect structure.

\paragraph{Effect of Foreground IoU Reward.}
Removing the foreground IoU reward raises the failure rate from 0.52\% to 9.87\% and reduces $\mathrm{IoU}_{\mathrm{fg}}$ from 0.2640 to 0.1795, demonstrating that this term complements rendering feedback by preserving the spatial extent of chart content.

\new{
\paragraph{Effect of CSG on Chart Editing.}
\label{para:csg_ablation}
Beyond reconstruction, we conduct a small controlled editing study to ablate the contribution of CSG to downstream manipulation. We select 100 charts from VisAnatomy to cover diverse chart types. For each chart, we obtain a CSG through LLM-assisted annotation followed by manual correction, yielding verified chart components and dependency edges. Each chart is paired with four manually created editing tasks—color update, axis rescaling, data value modification, and layout adjustment—together with manually edited reference SVGs. This study evaluates three aspects of the pipeline: whether CSG-guided editing improves success over direct SVG editing, whether DIVI-based semantic labels lead to accurate CSG construction, and whether the LLM uses the task-relevant CSG edges during edit propagation. All editing experiments use GPT-5.1 as the executor, while the CSG representation itself is model-agnostic and can be paired with other capable instruction-following models.
}

\input{tables/CSG_editing}

\new{
Table~\ref{tab:csg_editing} reports results across three configurations.
Without CSG, the LLM completes only 50.0\% of tasks.
Adding CSG-inferred constraints raises success rate to 63.0\% with edge propagation F1 of 0.5407, suggesting that explicit dependency edges help coordinate edits across related components.
When the heuristically tagged SVG is additionally provided as semantic context, all metrics improve further: success rate reaches 74.0\%, CSG construction accuracy reaches 89.2\%, and propagation F1 increases to 0.5782. These results indicate that both the CSG structure and the underlying semantic annotations contribute to more reliable chart editing, with semantic labels improving the construction and use of task-relevant dependency edges.
}

\paragraph{Summary.}
These ablation results support \new{three} key conclusions: (1) chart-specific SVG tokenization is crucial for learning structured chart syntax, (2) rendering-aware GRPO is a major driver of reconstruction quality, with the foreground IoU term providing additional gains in structural fidelity and output validity\new{, and (3) the CSG provides consistent, measurable benefit to downstream editing, validating it as a distinct and necessary contribution beyond reconstruction}.

%% file: tables/recon_id.tex
\begin{table*}[t]
\centering
\caption{In-distribution evaluation on Beagle+ test subsets.
↑: higher is better; ↓: lower is better. Bold: best among model-based methods; underline: second-best among model-based methods.}
\label{tab:indistribution}
\setlength{\tabcolsep}{3.2pt}
\renewcommand{\arraystretch}{1.15}
\resizebox{\textwidth}{!}{
\begin{tabular}{l|cccccc|cccccc|cccccc}
\toprule
\multirow{3}{*}{\textbf{Model}}
& \multicolumn{6}{c|}{\textbf{ChartBlocks Subset}}
& \multicolumn{6}{c|}{\textbf{Fusion Subset}}
& \multicolumn{6}{c}{\textbf{Plotly Subset}} \\
\cmidrule(lr){2-7} \cmidrule(lr){8-13} \cmidrule(l){14-19}
& \multirow{2}{*}{\textbf{PSNR} $\uparrow$}
& \multirow{2}{*}{\textbf{SSIM} $\uparrow$}
& \multirow{2}{*}{\textbf{LPIPS} $\downarrow$}
& \multirow{2}{*}{$\mathbf{IoU}_{\mathbf{fg}} \uparrow$}
& \multicolumn{1}{c}{\textbf{Edge}}
& \multicolumn{1}{c|}{\textbf{Failure}}
& \multirow{2}{*}{\textbf{PSNR} $\uparrow$}
& \multirow{2}{*}{\textbf{SSIM} $\uparrow$}
& \multirow{2}{*}{\textbf{LPIPS} $\downarrow$}
& \multirow{2}{*}{$\mathbf{IoU}_{\mathbf{fg}} \uparrow$}
& \multicolumn{1}{c}{\textbf{Edge}}
& \multicolumn{1}{c|}{\textbf{Failure}}
& \multirow{2}{*}{\textbf{PSNR} $\uparrow$}
& \multirow{2}{*}{\textbf{SSIM} $\uparrow$}
& \multirow{2}{*}{\textbf{LPIPS} $\downarrow$}
& \multirow{2}{*}{$\mathbf{IoU}_{\mathbf{fg}} \uparrow$}
& \multicolumn{1}{c}{\textbf{Edge}}
& \multicolumn{1}{c}{\textbf{Failure}} \\
&
&
&
&
& \multicolumn{1}{c}{\textbf{Cons.} $\uparrow$}
& \multicolumn{1}{c|}{\textbf{Rate} $\downarrow$}
&
&
&
&
& \multicolumn{1}{c}{\textbf{Cons.} $\uparrow$}
& \multicolumn{1}{c|}{\textbf{Rate} $\downarrow$}
&
&
&
&
& \multicolumn{1}{c}{\textbf{Cons.} $\uparrow$}
& \multicolumn{1}{c}{\textbf{Rate} $\downarrow$} \\

\rowcolor{gray!20}
Adobe Illustrator\textsuperscript{$\dagger$}
& \new{22.9034} & \new{0.9107} & \new{0.1736} & \new{0.9187} & \new{0.7276} & \new{0.0000}
& \new{21.3574} & \new{0.8813} & \new{0.2461} & \new{0.9257} & \new{0.5082} & \new{0.0000}
& \new{20.8586} & \new{0.8838} & \new{0.2174} & \new{0.9217} & \new{0.5202} & \new{0.0000} \\

\midrule
OmniSVG~\cite{yangomnisvg}
& 2.3979 & 0.1757 & 0.7943 & 0.2498 & 0.3705 & 0.2667
& 4.1156 & 0.2162 & 0.7501 & 0.3027 & 0.2814 & 0.1615
& 2.7260 & 0.0605 & 0.8878 & 0.1872 & 0.1359 & 0.5455 \\

StarVector~\cite{rodriguez2025starvector}
& 9.9510 & 0.4935 & 0.6870 & 0.4454 & 0.2497 & 0.2667
& 9.1777 & 0.4795 & 0.7316 & 0.3304 & 0.2161 & 0.1385
& 9.6040 & 0.4933 & 0.6955 & 0.2882 & 0.2240 & 0.1273 \\

ChartCoder~\cite{zhao2025chartcoder}
& 6.3438 & 0.4462 & 0.7897 & 0.1946 & 0.2872 & 0.1917
& 3.5305 & 0.2095 & 0.8778 & 0.2129 & 0.1008 & 0.5077
& 2.1856 & 0.1358 & 0.8838 & 0.2543 & 0.1172 & 0.4091 \\

GPT-4o
& 9.2612 & 0.5410 & 0.6406 & 0.4250 & 0.2825 & 0.0667
& 7.1797 & 0.4449 & 0.7012 & 0.3469 & 0.2018 & 0.1692
& 7.5700 & 0.3446 & 0.7571 & 0.2715 & 0.1479 & 0.3545 \\

o3
& \new{7.4921} & \new{0.4714} & \new{0.7072} & \new{0.2955} & \new{0.2808} & \new{0.1917}
& \new{\underline{10.5211}} & \new{\underline{0.5490}} & \new{0.6689} & \new{0.3537} & \new{0.2005} & \new{0.1308}
& \new{\underline{10.7894}} & \new{0.4870} & \new{0.6521} & \new{0.3481} & \new{0.1725} & \new{0.1545} \\

Gemini-2.5-Flash
& \new{9.9435} & \new{0.5984} & \new{0.6509} & \new{0.3733} & \new{0.3126} & \new{\underline{0.0167}}
& \new{\textbf{12.5728}} & \new{\textbf{0.6173}} & \new{0.5986} & \new{0.4388} & \new{0.2296} & \new{\underline{0.0077}}
& \new{\textbf{13.6669}} & \new{\textbf{0.6051}} & \new{\textbf{0.5520}} & \new{0.3938} & \new{0.2262} & \new{\underline{0.0182}} \\

\midrule
\ourmodel-4B
& \underline{15.1260} & \underline{0.7034} & \underline{0.2710} & \underline{0.6915} & \underline{0.4991} & \textbf{0}
& 8.7804 & 0.5020 & \textbf{0.5432} & \underline{0.5237} & \underline{0.3088} & \textbf{0}
& 7.4312 & \underline{0.4974} & 0.6099 & \textbf{0.4588} & \underline{0.2665} & \textbf{0} \\

\ourmodel-8B
& \textbf{15.5634} & \textbf{0.7106} & \textbf{0.2600} & \textbf{0.7075} & \textbf{0.5148} & \textbf{0}
& 8.7341 & 0.5111 & \underline{0.5569} & \textbf{0.5520} & \textbf{0.3140} & \textbf{0}
& 6.6080 & 0.4358 & \underline{0.6095} & \underline{0.4419} & \textbf{0.2895} & \textbf{0} \\

\bottomrule
\end{tabular}
}\vspace{-4mm}
\end{table*}

%% file: tables/recon_ood.tex
\begin{table}[t]
\centering
\caption{Out-of-distribution evaluation on VisAnatomy~\cite{chen2025visanatomy}. 
↑: higher is better; ↓: lower is better. Bold: best among model-based methods; underline: second-best among model-based methods.}
\label{tab:ood}
\setlength{\tabcolsep}{4.2pt}
\renewcommand{\arraystretch}{1.15}
\resizebox{\columnwidth}{!}{
\begin{tabular}{l|cccccc}
\toprule
\multirow{2}{*}{\textbf{Model}}
& \multirow{2}{*}{\textbf{PSNR} $\uparrow$}
& \multirow{2}{*}{\textbf{SSIM} $\uparrow$}
& \multirow{2}{*}{\textbf{LPIPS} $\downarrow$}
& \multirow{2}{*}{$\mathbf{IoU}_{\mathbf{fg}} \uparrow$}
& \multicolumn{1}{c}{\textbf{Edge}}
& \multicolumn{1}{c}{\textbf{Failure}} \\
& & & & & \multicolumn{1}{c}{\textbf{Cons.} $\uparrow$}
& \multicolumn{1}{c}{\textbf{Rate} $\downarrow$} \\

\rowcolor{gray!20}
Adobe Illustrator\textsuperscript{$\dagger$}
& \new{15.8100} & \new{0.7440} & \new{0.4726} & \new{0.3873} & \new{0.3691} & \new{0.0000} \\

\midrule
OmniSVG~\cite{yangomnisvg}
& 3.6713 & 0.2736 & 0.7947 & 0.1345 & 0.2755 & 0.2390 \\

StarVector~\cite{rodriguez2025starvector}
& 8.6618 & 0.4116 & 0.7714 & 0.1214 & 0.1685 & 0.3558 \\

ChartCoder~\cite{zhao2025chartcoder}
& 7.8008 & 0.4115 & 0.7898 & 0.0988 & 0.1864 & 0.4026 \\

GPT-4o
& 12.6482 & 0.6014 & 0.6066 & 0.2552 & 0.2423 & 0.0597 \\

o3
& \new{11.8976} & \new{0.5831} & \new{0.5808} & \new{\textbf{0.2821}} & \new{0.2551} & \new{0.1039} \\

Gemini-2.5-Flash
& \new{12.8705} & \new{0.6288} & \new{\underline{0.5669}} & \new{0.2667} & \new{0.2664} & \new{0.0364} \\

\midrule
\ourmodel-4B
& \underline{14.0211} & \underline{0.6951} & 0.5758 & 0.2640 & \underline{0.2919} & \textbf{0.0052} \\

\ourmodel-8B
& \textbf{14.2012} & \textbf{0.7021} & \textbf{0.5633} & \underline{0.2801} & \textbf{0.2992} & \underline{0.0078} \\

\bottomrule
\end{tabular}
}\vspace{-4mm}
\end{table}

%% file: tables/recon_data.tex
\begin{table}[t]
\centering
\caption{\new{Data accuracy by chart type on VisAnatomy~\cite{chen2025visanatomy}. For each 
chart, we retrieve the encoded data array and report the array-length matching rate (Len.) and the numerical accuracy over matched data (Acc.).}}\vspace{-2mm}
\label{tab:numerical_by_type_ood}
\setlength{\tabcolsep}{4.2pt}
\renewcommand{\arraystretch}{0.9}
\resizebox{0.9\columnwidth}{!}{
\begin{tabular}{l|cc|cc|cc}
\toprule
\multirow{2}{*}{\textbf{Model}}
& \multicolumn{2}{c|}{\textbf{Bar Chart}}
& \multicolumn{2}{c|}{\textbf{Pie Chart}}
& \multicolumn{2}{c}{\textbf{Line Chart}} \\
& \textbf{Len. $\uparrow$} & \textbf{Acc. $\uparrow$}
& \textbf{Len. $\uparrow$} & \textbf{Acc. $\uparrow$}
& \textbf{Len. $\uparrow$} & \textbf{Acc. $\uparrow$} \\
\midrule
StarVector~\cite{rodriguez2025starvector}
& \new{0} & \new{0} & \new{0} & \new{0} & \new{0} & \new{0} \\
OmniSVG~\cite{yangomnisvg}
& \new{0.0210} & \new{0} & \new{0.1039} & \new{0.0018} & \new{0} & \new{0} \\
GPT-4o
& \new{0.5944} & \new{0.0651} & \new{0.7013} & \new{0.0914} & \new{0.0870} & \new{0.0046} \\
o3
& \new{\textbf{0.7413}} & \new{0.1456} & \new{0.6623} & \new{0.1243} & \new{0.3913} & \new{0.1763} \\
Gemini-2.5-Flash
& \new{0.6643} & \new{0.1545} & \new{\textbf{0.7922}} & \new{\textbf{0.1664}} & \new{0.6087} & \new{0.2158} \\
\midrule
\ourmodel-8B
& \new{\underline{0.6713}} & \new{\textbf{0.1807}} & \new{\underline{0.7532}} & \new{\underline{0.1316}} & \new{\textbf{0.6522}} & \new{\textbf{0.2173}} \\
\bottomrule
\end{tabular}
}\vspace{-4mm}
\end{table}

%% file: tables/ablate_new.tex
\begin{table}[t]
\centering
\caption{Ablation studies on the 4B model. We take the full model as the reference and remove one component at a time. Higher is better for PSNR, SSIM, Foreground IoU ($\mathrm{IoU}_{\mathrm{fg}}$), and Edge Cons., while lower is better for LPIPS and Failure Rate.}\vspace{-2mm}
\label{tab:ablation}
\setlength{\tabcolsep}{4.2pt}
\renewcommand{\arraystretch}{1.12}
\resizebox{\columnwidth}{!}{
\begin{tabular}{l|cccccc}
\toprule
\multirow{2}{*}{\textbf{Variant}}
& \multirow{2}{*}{\textbf{PSNR} $\uparrow$}
& \multirow{2}{*}{\textbf{SSIM} $\uparrow$}
& \multirow{2}{*}{\textbf{LPIPS} $\downarrow$}
& \multirow{2}{*}{$\mathbf{IoU}_{\mathbf{fg}} \uparrow$}
& \multicolumn{1}{c}{\textbf{Edge}}
& \multicolumn{1}{c}{\textbf{Failure}} \\
& & & & & \multicolumn{1}{c}{\textbf{Cons.} $\uparrow$}
& \multicolumn{1}{c}{\textbf{Rate} $\downarrow$} \\
\midrule
\ourmodel-4B                & \textbf{14.0211} & \textbf{0.6951} & \textbf{0.5758} & 0.2640 & \textbf{0.2919} & \textbf{0.0052} \\
\midrule
w/o SVG tokens & 12.2325          & 0.6028          & 0.6358          & 0.1801 & 0.2585          & 0.1481 \\
w/o GRPO               & 6.8122           & 0.4672          & 0.6263          & \textbf{0.3771} & 0.2579          & 0.0260 \\
w/o $\mathrm{IoU}_{\mathrm{fg}}$ in GRPO        & 12.5594          & 0.6342          & 0.6255          & 0.1795 & 0.2616          & 0.0987 \\
\bottomrule
\end{tabular} 
}\vspace{-4mm}
\end{table}

%% file: tables/CSG_editing.tex

\begin{table}[t]
\centering
\caption{\new{Controlled editing study on 100 charts sampled from VisAnatomy. Each chart is paired with four manually specified editing tasks covering color, axis, data-value, and layout edits. CSG denotes whether CSG-inferred constraints are used during editing, and Semantic Label denotes whether the SVG is heuristically tagged before CSG construction. Success rate, construction accuracy, and edit-propagation F1 score are computed against manually verified annotations and reference edits.}}\vspace{-2mm}
\label{tab:csg_editing}
\setlength{\tabcolsep}{4pt}
\renewcommand{\arraystretch}{1.12}
\resizebox{\columnwidth}{!}{
\begin{tabular}{cc|ccccc}
\toprule

\new{\shortstack{\\\textbf{CSG}}} & \new{\shortstack{\textbf{Semantic}\\\textbf{Label}}} & \new{\shortstack{\textbf{Success}\\\textbf{Rate} $\uparrow$}} & \new{\shortstack{\textbf{CSG}\\\textbf{Acc.} $\uparrow$}} & 

\new{\shortstack{\textbf{Propagation}\\\textbf{Precision}\ $\uparrow$}} & \new{\shortstack{\textbf{Propagation}\\\textbf{Recall} $\uparrow$}} & \new{\shortstack{\textbf{Propagation}\\\textbf{F1} $\uparrow$}} \\

\midrule
\new{\ding{55}} & \new{\ding{55}} & \new{0.5000} & \new{---}    & \new{---}    & \new{---}    & \new{---} \\
\new{\ding{51}} & \new{\ding{55}} & \new{0.6300} & \new{0.8389} & \new{0.6194} & \new{0.4798} & \new{0.5407} \\
\new{\ding{51}} & \new{\ding{51}} & \new{\textbf{0.7400}} & \new{\textbf{0.8919}} & \new{\textbf{0.7025}} & \new{\textbf{0.4913}} & \new{\textbf{0.5782}} \\
\bottomrule
\end{tabular}
}\vspace{-4mm}
\end{table}

%% file: sections/6_case-study.tex
\section{Case Study} \label{sec:case_study}

By transforming raster chart images into manipulable semantic representations, Chart2SVG enables three types of language-driven interactions: \textit{interactive exploration}, \textit{chart repurposing}, and \textit{layout reuse}. In all experiments, Chart2SVG first reconstructs the chart SVG and semantic structure from the raster image, after which GPT-5.1 is used solely downstream to interpret user instructions and generate editing actions over the reconstructed representation.

\begin{figure}[t]
  \centering
  \includegraphics[width=0.5\textwidth]{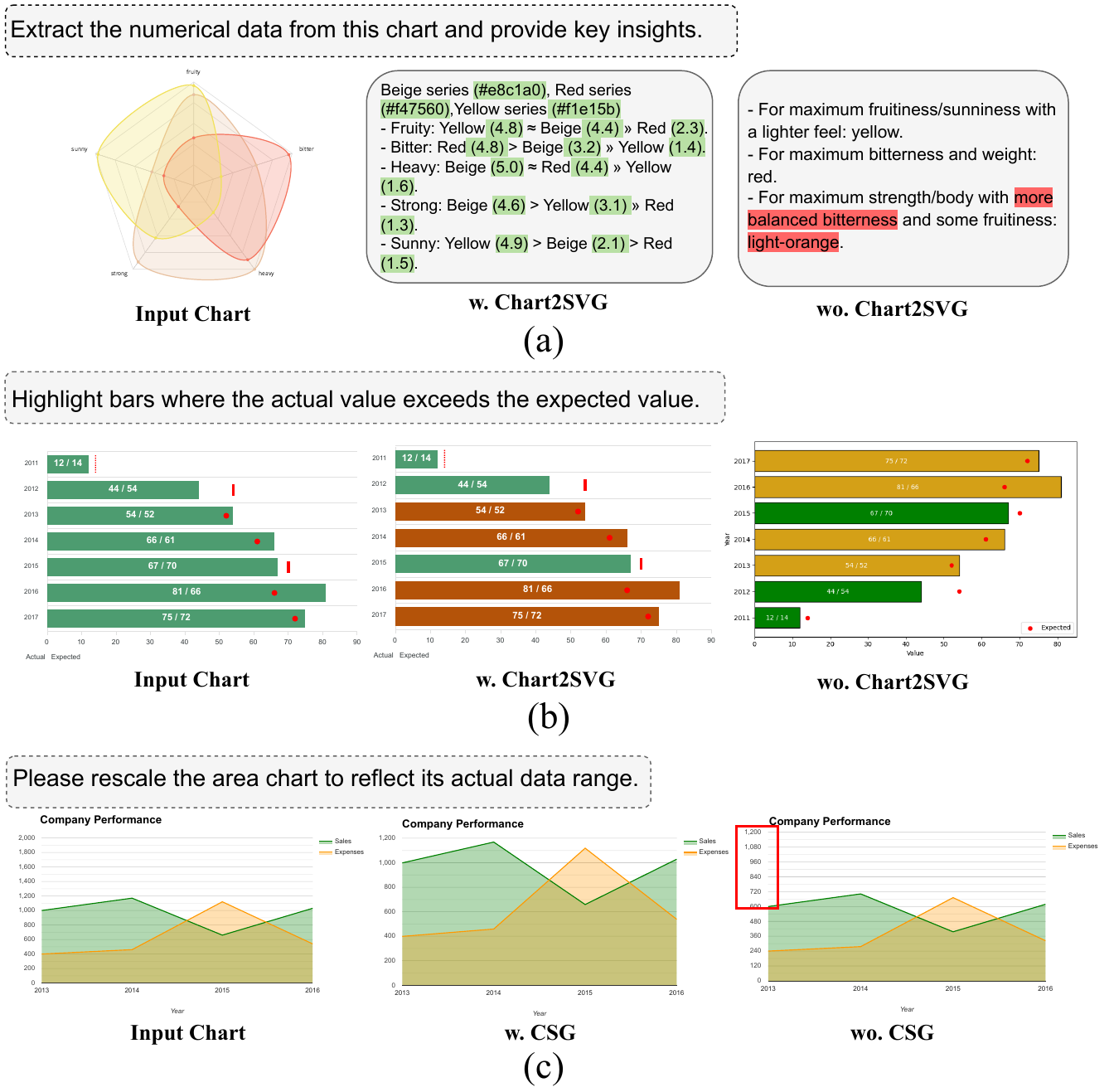}
  \vspace{-6mm}
\caption{Interactive chart exploration. (a) Data extraction: Chart2SVG enables precise value recovery; baseline hallucinates ranges. (b) Semantic querying: correct mark highlighting vs. baseline 
misidentification. (c) Rescaling: CSG updates geometric paths; baseline only edits text labels.}
  \label{fig:data_analysis}\vspace{-6mm}
\end{figure}

\paragraph{Interactive Exploration.}
\ourmodel transforms chart interaction from passive viewing into active visual exploration by allowing the LLM to operate over structured chart representations rather than raw pixels. To evaluate this, we compare tasks in Fig.~\ref{fig:data_analysis}(a) and (b) against the same LLM applied directly to raster images.

In Fig.~\ref{fig:data_analysis}(a), access to the reconstructed SVG enables the LLM to recover exact numerical values, such as minimums and peak totals, even when explicit labels are absent. In contrast, when operating on raster inputs, the LLM must rely on visual estimation, leading to vague approximations and hallucinated ranges (highlighted in red).

Beyond analysis, our approach supports semantic querying by treating visual marks as structured, addressable entities. In the conditional highlighting task (Fig.~\ref{fig:data_analysis}(b)), the LLM accurately identifies and highlights bars where actual values exceed expected values. Without the structured representation, the raster baseline fails to follow this logic and misidentifies the relevant data points.

To isolate the role of explicit structure, Fig.~\ref{fig:data_analysis}(c) compares edits generated with and without the CSG. When asked to rescale the y-axis to the true data range, the LLM leverages the dependencies encoded in the CSG to update geometric paths and maintain consistency between axes and marks. Without the CSG, the model only modifies textual labels, leaving the underlying geometry unchanged and breaking the visual-data correspondence.

\begin{figure}[t]
  \centering
  \includegraphics[width=0.5\textwidth]{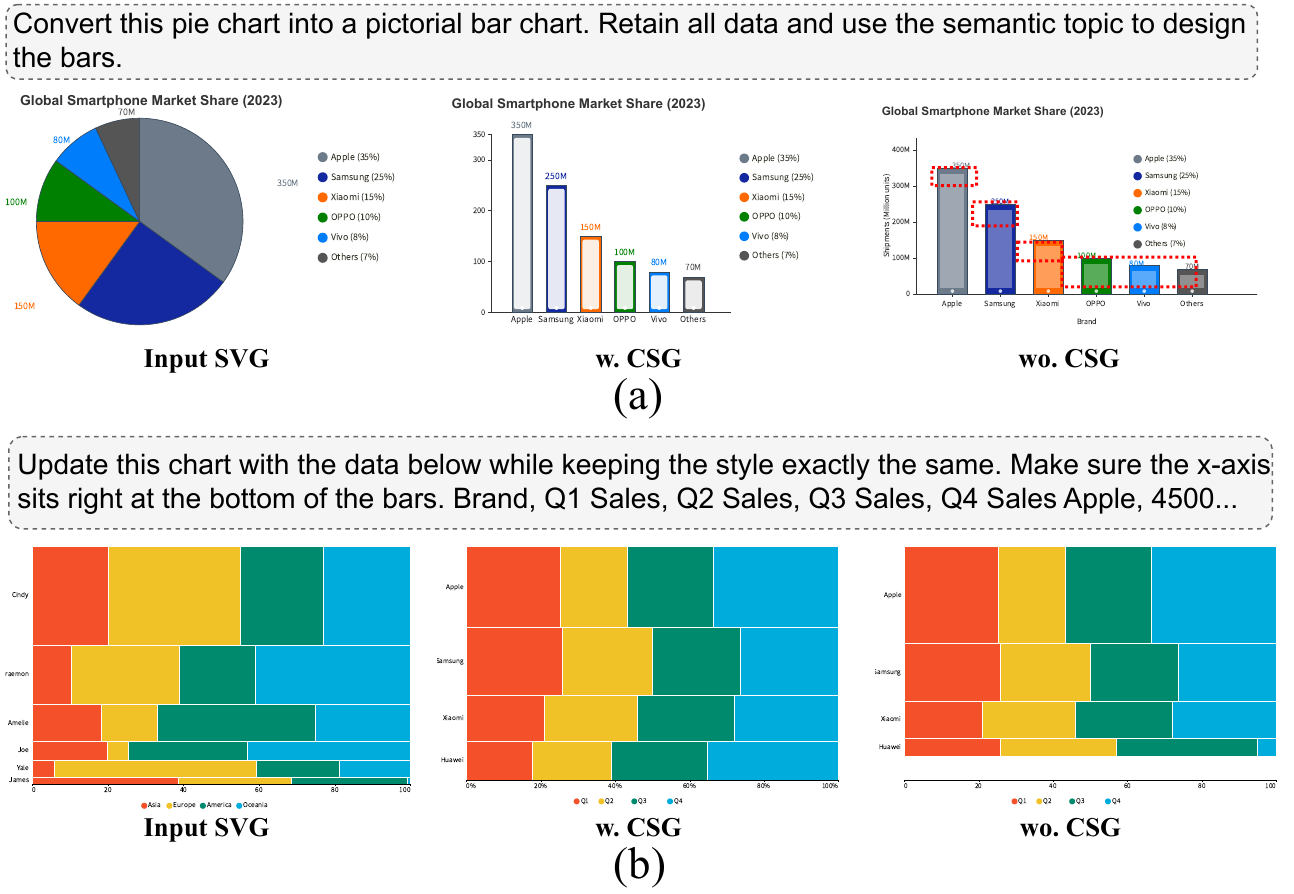}  
  \vspace{-6mm}
  \caption{Chart repurposing (a) and layout reuse (b). (a) CSG enables pie-to-pictorial-bar conversion; baseline hallucinates data (red). (b) CSG preserves x-axis alignment under new data; baseline fails.}
   \vspace{-6mm}
  \label{fig:repurpose}
\end{figure}

\paragraph{Chart Repurposing.}
Charts can be transformed to re-encode existing data into new visual forms for more effective communication.
In Fig.~\ref{fig:repurpose}(a), a pie chart is converted into a pictorial bar chart. Guided by CSG dependencies, the model preserves numerical fidelity while adapting the visual form; without the CSG, the LLM reverts to generic geometries and hallucinates data (red).

\paragraph{Layout Reuse.}
Existing chart layouts can also be reused to visualize new datasets while preserving the original design. Fig.~\ref{fig:repurpose}(b) shows a normalized stacked bar chart adapted to new sales data.The CSG enforces key spatial constraints, such as aligning the x-axis with the bar bases, ensuring structural consistency. Without it, the LLM introduces gaps that break the layout and degrade visual fdelity.

%% file: sections/6_disscusion.tex
\section{Discussion and Conclusion}
We introduced \ourmodel, a framework that transforms static raster 
charts into \new{semantically structured SVG that enable 
programmatic and language-guided editing}. By 
combining specialized training objectives with a rendering-aware 
post-training phase, \ourmodel produces charts that are 
visually accurate and structurally organized for direct manipulation. 
Furthermore, by constructing and integrating a chart structure graph, 
we make latent visual dependencies explicit, enabling complex tasks 
such as interactive data exploration and automated repurposing.

\paragraph{Generalizability.}
While \ourmodel performs well on common chart types, its capabilities reflect the Beagle+ dataset, which is dominated by foundational marks such as rectangles, lines, and circles, with complex marks accounting for only $\approx$5\% of the corpus.
\new{The training set also excludes the D3 subset of Beagle and SVGs beyond the 8192-token limit, leaving high-density or highly customized charts underrepresented. Consequently, generalization to irregular layouts, degraded screenshots, and unconventional mark types remains limited. Future work will augment training with synthetic data covering complex marks.}

\paragraph{Data Fidelity.}
\new{\ourmodel prioritizes visual fidelity and structural editability over explicit recovery of the underlying data table. Consequently, precise data binding—such as axis-scale inference, category-to-mark alignment, and value-to-geometry mapping—remains an important direction for future work. Integrating OCR, chart-to-table extraction, and scale-fitting modules with the reconstructed SVG could further improve data-level correctness while preserving editability.}

\paragraph{Broader Applications.}
Like any deep learning model, Chart2SVG cannot guarantee perfect accuracy on degraded or custom charts, making human-in-the-loop refinement essential. Future work could integrate Chart2SVG with formal chart grammars for rule-based refinement, or connect the queryable CSG with autonomous AI agents for complex analytical queries and multi-chart synthesis.

%% file: sections/7_Ack.tex
\section*{Acknowledgment}
We sincerely thank Xuan Hua and Hewen Zhang for their experiments and data generation support. This work was supported by the grants of NSFC (No.U2436209), the Beijing Natural Science Foundation (L247027), the Fundamental Research Funds for the Central Universities, and the Research Funds of Renmin University of China.

%% file: appendix.tex
\appendix
\label{sec:appendix}

\section{Full Definition of SVG-Specific Tokens}

As introduced in Section 3.2 of the main text, representing SVG charts as raw character-level XML strings is inefficient and highly susceptible to syntactic hallucinations. To enable the Vision-Language Model (VLM) to better capture the semantic and hierarchical structure of charts, we extend the tokenizer vocabulary with 115 SVG-specific special tokens. 

This tokenization strategy not only significantly compresses the sequence length but also forces the model to learn chart-oriented vector representations rather than unstructured text. As detailed in Table \ref{tab:svg_tokens_wide}, these 115 tokens are systematically categorized into four functional groups based on their roles in chart rendering and data encoding:

\begin{itemize}
    \item \textbf{1. Structural Containers \& Anchors:} This group includes paired start and end tags for document scopes, groupings (e.g., \texttt{<g>}), definitions (\texttt{<defs>}), and advanced rendering blocks (e.g., gradients and filters). These tokens are critical for maintaining the hierarchical DOM structure of the chart, allowing multiple primitive marks to be logically grouped (e.g., grouping bars of the same data series) to support downstream Chart Structure Graph (CSG) extraction and layout reuse.
    
    \item \textbf{2. Graphical Primitives:} This group covers the fundamental vector shapes utilized to represent data marks, including basic geometric shapes (\texttt{<rect>}, \texttt{<circle>}, \texttt{<line>}), complex trajectories (\texttt{<path>}), and typography (\texttt{<text>}, \texttt{<tspan>}). By explicitly tokenizing their opening and closing tags, the model is guided to generate strictly bounded and syntactically valid primitive components.
    
    \item \textbf{3. Geometric Opcodes:} To accurately reconstruct highly customized shapes and continuous data boundaries (e.g., in area charts or radar charts), we tokenize the low-level path commands. We explicitly differentiate between absolute (\texttt{abs}) and relative (\texttt{rel}) coordinate instructions (e.g., \texttt{moveto}, \texttt{lineto}, \texttt{curveto}). This allows the model to precisely execute complex trajectory planning without being distracted by raw character parsing.
    
    \item \textbf{4. Visual Channel \& Geometry Attributes:} This group encapsulates the attributes that map underlying data to visual representations. It includes spatial coordinates and dimensions (\texttt{x=}, \texttt{width=}), color and styling aesthetics (\texttt{fill=}, \texttt{stroke=}), typographic properties (\texttt{font-size=}, \texttt{text-anchor=}), and affine transformations (\texttt{transform=}). Formulating these properties as discrete tokens enables the model to disentangle visual styling from geometric positioning.
\end{itemize}

The complete list of the 115 SVG-specific special tokens and their corresponding descriptions is provided in Table \ref{tab:svg_tokens_wide}.

\vspace{1em}
\section{Prompt Design for Chart Structure Graph Construction and Manipulation}
\label{sec:CSG_prompt}

To enable language-guided chart manipulation, our framework employs a two-stage prompting strategy. First, the LLM analyzes the parsed SVG to extract a Scene Graph, explicitly inferring topological dependencies and structural hierarchies. Second, leveraging this graph and the raw SVG code, the LLM executes downstream user requests—dynamically switching between an \textit{Analysis Mode} (for numerical insights) and an \textit{Editing Mode} (for visual modifications). 

Both prompts are carefully engineered to enforce strict structural preservation. By assigning an expert persona and defining mandatory constraint resolution steps (e.g., coordinate shifting and canvas expansion), we prevent visual artifacts and ensure the edited SVGs remain valid and geometrically consistent. The complete templates are provided below:

\vspace{1em}

\begin{tcolorbox}[
    colback=gray!5!white,      
    colframe=gray!50!black,    
    title=\textbf{System Prompt for Scene Graph Extraction}, 
    fonttitle=\bfseries\small,
    boxrule=0.5pt,             
    arc=2pt,                   
    left=4pt, right=4pt, top=4pt, bottom=4pt 
]
\small
\textbf{[System Role]} \\
You are an expert SVG scene graph generator.

\vspace{0.5em}
You are given an SVG string and a Scene Graph JSON template. \\
Your task is to analyze the SVG and populate the template with concrete values.

\vspace{0.5em}
\textbf{Requirements:} \\
1. Strictly follow the provided JSON schema. Do NOT change field names or structure. \\
2. You MAY expand list-type fields (e.g., marks, ticks, legend items) to include multiple instances if needed. \\
3. Do NOT add new keys that are not defined in the templates. \\
4. Only fill in values that can be directly inferred from the SVG. \\
5. If a value cannot be determined, set it to null. \\
6. Preserve all hierarchical relationships and nesting.

\vspace{0.5em}
\textbf{Output Format:} \\
- Return ONLY a valid JSON object for the scene graph. \\
- Do NOT include any explanations, comments, or extra text.
\end{tcolorbox}




\vspace{1em}

\begin{tcolorbox}[
    colback=gray!5!white,      
    colframe=gray!50!black,    
    title=\textbf{System Prompt for CSG-Guided Chart Reasoning and Manipulation}, 
    fonttitle=\bfseries\small,
    boxrule=0.5pt,             
    arc=2pt,                   
    left=4pt, right=4pt, top=4pt, bottom=4pt 
]
\small
\textbf{[System Role]} \\
You are an expert SVG editor.

\vspace{0.5em}
\textbf{[If \texttt{analysis\_mode} is True:]} \\
You are given: \\
1. An SVG chart \\
2. The dependency schema representing the relationships between elements \\
3. A user question 

\textbf{\#\# Task} \\
Answer the question by analyzing the SVG and dependencies. Provide a clear, direct answer. 

\textbf{\#\# Output} \\
Return plain text only. Do not output SVG. 

\vspace{0.8em}
\hrule
\vspace{0.8em}

\textbf{[If \texttt{analysis\_mode} is False (Editing Mode):]} \\
You are given: \\
1. An SVG chart \\
2. The dependency schema representing the relationships between elements \\
3. An editing instruction 

\textbf{\#\# Task} \\
Perform a \textbf{global, structure-aware edit} of the SVG. 

---

\textbf{\#\# Process (MANDATORY)} \\
1. Identify directly affected elements and their original values according to the svg code and scene graph. \\
2. Trace dependencies using the provided JSONs (Pay special attention to \texttt{layout\_cascade}). \\
3. CALCULATE the new quantitative values. If mark width/height changes and padding is fixed, mathematically deduce the new axis length and viewBox dimensions. \\
4. Apply edits in a consistent order: element $\rightarrow$ layout $\rightarrow$ axis $\rightarrow$ dependent elements. \\
Do not skip dependency reasoning. 

\textbf{\#\# CRITICAL CONSTRAINTS (DO NOT VIOLATE)} \\
1. \textbf{ZERO DELETION}: You MUST preserve EVERY single element from the original SVG (axes, text labels, legends, ALL segments and colors of stacked bars). Do NOT delete any nodes unless explicitly requested. \\
2. \textbf{NO SHORTCUTS}: Do NOT use comments like \texttt{\textless{}!-- rest of the code --\textgreater{}} or \texttt{...}. You MUST output the complete, fully renderable SVG code. \\
3. \textbf{COORDINATE SHIFTING}: If you change the thickness/size of an element, you MUST mathematically recalculate and shift the position (e.g., \texttt{x} or \texttt{y} attribute) of ALL subsequent elements and groups. You must manually add the size difference to their coordinates to maintain the original padding/gap. Elements MUST NOT overlap. \\
4. \textbf{CANVAS EXPANSION}: If the total computed layout size increases, you MUST expand the \texttt{\textless{}svg viewBox="..."\textgreater{}} and axis lines accordingly so nothing is clipped. 

\textbf{\#\# Requirements} \\
1. Preserve stacking, alignment, and spacing constraints. \\
2. Maintain consistent gaps and proportions. \\
3. Ensure all elements remain within valid axis ranges. 

\textbf{\#\# Output Format (STRICT)} \\
Output the COMPLETE modified SVG inside an \texttt{```xml} code block. 

\textbf{\#\# Goal} \\
Modify the SVG while preserving all structural relationships defined by the provided JSONs.
\end{tcolorbox}



\newcolumntype{W}{>{\raggedright\arraybackslash\let\oldunderscore\_\renewcommand{\_}{\oldunderscore\allowbreak}}X}
\newcolumntype{Y}{>{\raggedright\arraybackslash}X}

\begin{table*}[t]
\centering
\scriptsize 
\setlength{\tabcolsep}{3pt} 
\renewcommand{\arraystretch}{1.1}

\caption{Full Definition of the 115 SVG-Specific Tokens}
\label{tab:svg_tokens_wide}

\begin{tabularx}{\textwidth}{@{} W Y | W Y @{}}
\toprule
\textbf{Token} & \textbf{Description} & \textbf{Token} & \textbf{Description} \\
\midrule

\multicolumn{4}{l}{\textbf{1. Structural Containers \& Anchors}} \\
\midrule
\texttt{[<|START\_OF\_SVG|>]} / \texttt{[<|END\_OF\_SVG|>]} & SVG start/end &
\texttt{[<|START\_OF\_GROUP|>]} / \texttt{[<|END\_OF\_GROUP|>]} & Group (\texttt{<g>}) start/end \\

\texttt{[<|clipPath|>]} / \texttt{[<|/clipPath|>]} & Clipping path start/end &
\texttt{[<|defs|>]} / \texttt{[<|/defs|>]} & Reusable elements start/end \\

\texttt{[<|linearGradient|>]} / \texttt{[<|/linearGradient|>]} & Linear gradient start/end &
\texttt{[<|radialGradient|>]} / \texttt{[<|/radialGradient|>]} & Radial gradient start/end \\

\texttt{[<|stop|>]} / \texttt{[<|/stop|>]} & Gradient stop start/end &
\texttt{[<|filter|>]} / \texttt{[<|/filter|>]} & Filter element start/end \\

\texttt{[<|feOffset|>]} / \texttt{[<|/feOffset|>]} & feOffset start/end &
\texttt{[<|feColorMatrix|>]} / \texttt{[<|/feColorMatrix|>]} & feColorMatrix start/end \\

\texttt{[<|feGaussianBlur|>]} / \texttt{[<|/feGaussianBlur|>]} & feGaussianBlur start/end &
\texttt{[<|feComposite|>]} / \texttt{[<|/feComposite|>]} & feComposite start/end \\

\midrule

\multicolumn{4}{l}{\textbf{2. Graphical Primitives}} \\
\midrule
\texttt{[<|path|>]} / \texttt{[<|/path|>]} & Path start/end &
\texttt{[<|circle|>]} / \texttt{[<|/circle|>]} & Circle start/end \\

\texttt{[<|rect|>]} / \texttt{[<|/rect|>]} & Rectangle start/end &
\texttt{[<|ellipse|>]} / \texttt{[<|/ellipse|>]} & Ellipse start/end \\

\texttt{[<|polygon|>]} / \texttt{[<|/polygon|>]} & Polygon start/end &
\texttt{[<|line|>]} / \texttt{[<|/line|>]} & Line start/end \\

\texttt{[<|polyline|>]} / \texttt{[<|/polyline|>]} & Polyline start/end &
\texttt{[<|text|>]} / \texttt{[<|/text|>]} & Text start/end \\

\texttt{[<|tspan|>]} / \texttt{[<|/tspan|>]} & tspan start/end &
\texttt{[<|use|>]} / \texttt{[<|/use|>]} & Uses object start/end \\

\midrule

\multicolumn{4}{l}{\textbf{3. Geometric Opcodes}} \\
\midrule
\texttt{[<|moveto\_abs|>]} / \texttt{[<|moveto\_rel|>]} & Move abs/rel &
\texttt{[<|lineto\_abs|>]} / \texttt{[<|lineto\_rel|>]} & Line abs/rel \\

\texttt{[<|horizontal\_lineto\_abs|>]} / \texttt{[<|horizontal\_lineto\_rel|>]} & H line abs/rel &
\texttt{[<|vertical\_lineto\_abs|>]} / \texttt{[<|vertical\_lineto\_rel|>]} & V line abs/rel \\

\texttt{[<|curveto\_abs|>]} / \texttt{[<|curveto\_rel|>]} & Cubic Bézier abs/rel &
\texttt{[<|smooth\_curveto\_abs|>]} / \texttt{[<|smooth\_curveto\_rel|>]} & Smooth cubic abs/rel \\

\texttt{[<|quadratic\_bezier\_curve\_abs|>]} / \texttt{[<|quadratic\_bezier\_curve\_rel|>]} & Quad Bézier abs/rel &
\texttt{[<|smooth\_quadratic\_bezier\_curveto\_abs|>]} / \texttt{[<|smooth\_quadratic\_bezier\_curveto\_rel|>]} & Smooth quad abs/rel \\

\texttt{[<|elliptical\_arc\_abs|>]} / \texttt{[<|elliptical\_arc\_rel|>]} & Arc abs/rel &
\texttt{[<|close\_path|>]} & Close path \\

\midrule

\multicolumn{4}{l}{\textbf{4. Visual Channel \& Geometry Attributes}} \\
\midrule
\texttt{[<|id=|>]} & ID &
\texttt{[<|class=|>]} & Class \\

\texttt{[<|d=|>]} & Path data &
\texttt{[<|style=|>]} & Style \\

\texttt{[<|fill=|>]} & Fill color &
\texttt{[<|fill-opacity=|>]} & Fill opacity \\

\texttt{[<|stroke=|>]} & Stroke color &
\texttt{[<|stroke-width=|>]} & Stroke width \\

\texttt{[<|stroke-linecap=|>]} & Line cap &
\texttt{[<|stroke-opacity=|>]} & Stroke opacity \\

\texttt{[<|stroke-dasharray=|>]} & Dash pattern &
\texttt{[<|opacity=|>]} & Global opacity \\

\texttt{[<|transform=|>]} & Transform &
\texttt{[<|gradientTransform=|>]} & Gradient transform \\

\texttt{[<|width=|>]} & Element width &
\texttt{[<|height=|>]} & Element height \\

\texttt{[<|x=|>]} & Position x &
\texttt{[<|y=|>]} & Position y \\

\texttt{[<|dx=|>]} & Shift x &
\texttt{[<|dy=|>]} & Shift y \\

\texttt{[<|cx=|>]} & Center x &
\texttt{[<|cy=|>]} & Center y \\

\texttt{[<|rx=|>]} & Radius x &
\texttt{[<|ry=|>]} & Radius y \\

\texttt{[<|r=|>]} & Circle radius &
\texttt{[<|points=|>]} & Polygon points \\

\texttt{[<|x1=|>]} & Start x &
\texttt{[<|y1=|>]} & Start y \\

\texttt{[<|x2=|>]} & End x &
\texttt{[<|y2=|>]} & End y \\

\texttt{[<|offset=|>]} & Gradient offset &
\texttt{[<|stop-color=|>]} & Stop color \\

\texttt{[<|stop-opacity=|>]} & Stop opacity &
\texttt{[<|href=|>]} & External reference \\

\texttt{[<|fr=|>]} & Radial focal r &
\texttt{[<|fx=|>]} & Radial focal x \\

\texttt{[<|fy=|>]} & Radial focal y &
\texttt{[<|filter=|>]} & Filter attribute \\

\texttt{[<|in=|>]} & Filter input &
\texttt{[<|stdDeviation=|>]} & Gaussian blur std dev \\

\texttt{[<|rotate=|>]} & Rotation angle &
\texttt{[<|font-size=|>]} & Font size \\

\texttt{[<|font-style=|>]} & Font style &
\texttt{[<|font-family=|>]} & Font family \\

\texttt{[<|font-weight=|>]} & Font weight &
\texttt{[<|text-anchor=|>]} & Text anchor \\

\texttt{[<|text-content=|>]} & Text content &
\texttt{[<|dominant-baseline=|>]} & Dominant baseline \\

\texttt{[<|viewBox=|>]} & ViewBox &
\texttt{[<|clip-path=|>]} & Clip path attr \\

\texttt{[<|display=|>]} & Display &
\texttt{[<|visibility=|>]} & Visibility \\

\bottomrule
\end{tabularx}
\end{table*}

\begin{figure*}[t]
    \centering
    \includegraphics[width=\linewidth]{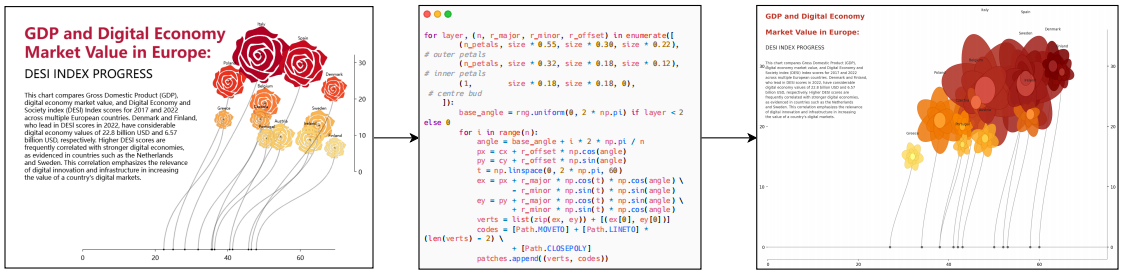}
    \caption{\new{Failure case of chart-to-code methods on a highly customized infographic. (Left) Original raster chart using rose-shaped glyphs to jointly encode GDP by size, DESI score by color, and country ranking by position. (Center) A chart-to-code approach attempts to reproduce the customized rose marks with
    Matplotlib; because the library lacks native support for such pictorial glyphs, the petals must be manually approximated using parameterized ellipses. (Right) The resulting rendering captures only the coarse layout and data encoding, while failing to preserve the distinctive rose-petal geometry and visual style of the original chart. This example illustrates the difficulty of
    chart-to-code methods in reconstructing highly customized
    pictorial marks with predefined plotting primitives.}}
    \label{fig:rose_case}
\end{figure*}

\begin{figure*}[t]
    \centering
    \includegraphics[width=\linewidth]{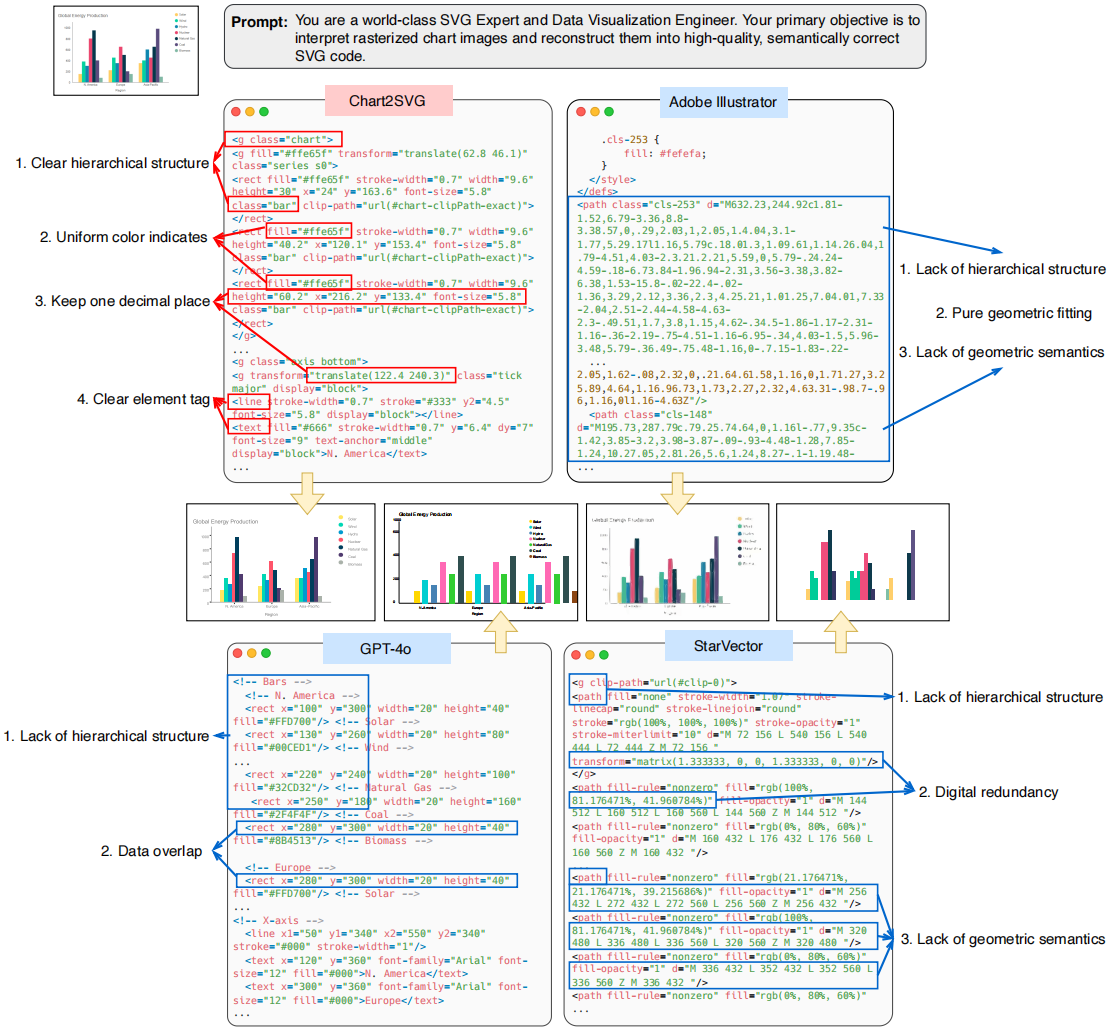}
    \caption{\new{Qualitative comparison of SVG code structure across 
    different methods on a grouped bar chart. 
    Chart2SVG produces a clear hierarchical structure and uniform color and coordinate representations, enabling direct programmatic editing. 
    Adobe Illustrator Image Trace generates a flat list of anonymous \texttt{<path>} elements with opaque Bézier sequences and no semantic structure. 
    GPT-4o produces partially readable code but lacks proper grouping and contains data overlaps. 
    StarVector outputs path-based geometry with numerical redundancy and no semantic labels.}}
    \label{fig:svg_code_case}
\end{figure*}

\section{\new{GRPO IMPLEMENTATION DETAILS}}
\label{grpo}

\subsection*{\new{Reward Design}}

\new{All three visual reward terms are normalized to $[0,1]$ before aggregation to ensure balanced contribution and to prevent any single term from dominating the combined reward due to differing numerical scales.}

\new{\paragraph{Foreground IoU.}
For $\mathrm{IoU_{fg}}$, we apply a relaxed foreground matching strategy to tolerate minor spatial misalignments common in autoregressive SVG generation. Foreground pixels are identified as those whose grayscale value falls below a threshold of 245 (on a 0--255 scale), separating chart content from the background. The predicted foreground mask $\hat{M}$ is dilated with a $5{\times}5$ structuring element prior to intersection computation, while the union is computed from the original undilated masks:}
\begin{equation}
    \new{\mathrm{IoU_{fg}}(\tilde{I}_i, I) = 
    \frac{|\mathrm{dilate}(\hat{M}) \cap M|}{|\hat{M} \cup M| + \varepsilon},}
\end{equation}
\new{where $\varepsilon$ is a small constant preventing division by zero, and the final score is clipped to $[0,1]$ to handle the asymmetric dilation that may cause the numerator to exceed the denominator. The dilation tolerates
sub-pixel rendering offsets between the predicted and ground-truth SVG without penalizing structurally correct but slightly misaligned reconstructions.}

\new{\paragraph{PSNR.}
For $\mathrm{PSNR}$, the raw decibel value is normalized by a ceiling of
40\,dB, which encompasses the practical reconstruction quality range
observed across all models in our setting (typically 2--16\,dB, as
reported in Tables~1--2 of the main text):}
\begin{equation}
    \new{\mathrm{PSNR}(\tilde{I}_i, I) =
    \min\!\left(
        \frac{20\log_{10}(255) - 10\log_{10}(\mathrm{MSE} + \varepsilon)}
             {40},\ 1
    \right),}
\end{equation}
\new{where $\mathrm{MSE}$ is the mean squared pixel error between $\tilde{I}_i$ and $I$, and $\varepsilon$ handles the degenerate case of near-perfect reconstruction. In practice, $\mathrm{PSNR_{norm}}$ operates in $[0.05, 0.39]$ for our models, a range comparable to or lower than
the $\mathrm{IoU_{fg}}$ and $\mathrm{SSIM}$ terms, confirming that PSNR does not dominate the combined reward despite its unbounded raw scale. Note that this normalization applies only within the GRPO visual 
reward; the PSNR reported in Tables~1--2 uses the standard 
unbounded decibel scale for comparability with prior work.}

\new{\paragraph{SSIM.}
For $\mathrm{SSIM}$, we compute a global structural similarity over
Gaussian-smoothed ($\sigma{=}1.5$) grayscale images:}
\begin{equation}
    \new{\mathrm{SSIM}(\tilde{I}_i, I) =
    \frac{2\mu_x\mu_y + C_1}{\mu_x^2 + \mu_y^2 + C_1}
    \cdot
    \frac{2|\sigma_{xy}| + C_2}{\sigma_x^2 + \sigma_y^2 + C_2},}
\end{equation}
\new{where $\mu_x, \mu_y$ and $\sigma_x^2, \sigma_y^2$ denote the global
mean and variance of the two Gaussian-smoothed images, $\sigma_{xy}$
their cross-covariance, $C_1{=}(0.01{\times}255)^2$, and
$C_2{=}(0.03{\times}255)^2$. Unlike the standard SSIM formulation, which
uses the signed cross-covariance $\sigma_{xy}$, we take its absolute
value $|\sigma_{xy}|$ to ensure the reward remains non-negative even
when the predicted and ground-truth images are weakly anti-correlated
in local structure, avoiding spurious negative reward signals during
early training.}

\new{\paragraph{Combined Reward.}
The three normalized terms are averaged uniformly to form the visual
reward, as in Eq.~(6) of the main text:}
\begin{equation}
    \new{R_{\mathrm{vis}}(\tilde{I}_i, I) = \frac{1}{3}\Bigl(
        \mathrm{IoU_{fg}}(\tilde{I}_i, I) +
        \mathrm{PSNR_{norm}}(\tilde{I}_i, I) +
        \mathrm{SSIM}(\tilde{I}_i, I)
    \Bigr).}
\end{equation}
\new{This combined visual reward is further weighted against the
code-level reward $R_{\mathrm{code}}$ as described in Eq.~(7), with
$\lambda_{\mathrm{vis}} = 0.6$ and $\lambda_{\mathrm{code}} = 0.2$ (Section\ref{sec:eval}), ensuring that optimization prioritizes faithful visual reconstruction while still penalizing invalid or excessively verbose SVG outputs.}

\subsection*{\new{Relation to Domain-general Visual RL}}

\new{Our reward design differs from domain-general visual RL
approaches such as MSRL~\cite{wang2026msrl} and Reason-SVG~\cite{xing2025reason} in two key respects.
First, both MSRL and Reason-SVG target code generation or general illustration SVG synthesis, where computing perceptual similarity over the full canvas is appropriate; for chart images, the foreground-aware $\mathrm{IoU_{fg}}$ is necessary to prevent background regions from dominating the reward signal. Second, MSRL uses execution-based rewards that require access to ground-truth data tables, whereas our rewards operate entirely in pixel space, requiring no auxiliary data source beyond the input raster image itself.}

\section{\new{Baseline Settings and Comparative Analysis}}
\label{baseline_setting}

\new{
\paragraph{Implementation Details.}
All baselines use officially released checkpoints without any fine-tuning on Beagle+ or chart-specific data, and are evaluated in a zero-shot manner under a unified prompt (shown in Fig.~\ref{fig:svg_code_case}).
Input images are resized to $512{\times}512$ for all models.
Since autoregressive generation introduces stochasticity, all results are averaged over four independent runs per sample.
We set the maximum output length to 8192 tokens across all models to ensure complete generation without truncation.
StarVector-8B is evaluated with temperature$=1.5$, 
repetition\_penalty$=3.1$; OmniSVG-3B with temperature$=0.3$, top\_p$=0.90$, repetition\_penalty$=1.05$; and Chart2SVG with temperature$=0.75$, repetition\_penalty$=1.05$.
ChartCoder generates executable Matplotlib/Seaborn code rather than SVG; 
we execute the output and save the rendered image for evaluation. GPT-4o is queried with temperature$=0$ for deterministic outputs and its SVG is parsed directly without post-hoc rendering repair.}

\new{\paragraph{Analysis of Baseline Performance.}
StarVector and OmniSVG perform substantially worse on chart reconstruction for two compounding reasons.
First, both models are trained predominantly on icon, emoji, and illustration datasets~\cite{rodriguez2025starvector, yangomnisvg}, with no chart-specific training data; their learned priors are poorly suited to the structured geometric regularity of visualization charts.
Second, both models represent all geometric shapes exclusively via \texttt{<path>} commands with Bézier sequences, even for primitives such as rectangles and lines that can be expressed far more compactly as \texttt{<rect>} or \texttt{<line>} elements.
This path-based representation consumes substantially more tokens per visual element than semantic primitives: a single bar in a bar chart, for instance, requires a closed five-point path (\texttt{M...L...L...Z}) instead of a single \texttt{<rect>} tag.
Chart SVGs are inherently more complex than icons or emoji---comprising axes, ticks, gridlines, legends, and multi-series marks---whereas typical icon SVGs can be fully expressed within 1,024 tokens.
The combination of domain mismatch and token-inefficient representation makes faithful chart reconstruction beyond the effective capacity of these models, as reflected in their high failure rates and fragmented outputs in Table~\ref{tab:indistribution} and Fig.~\ref{fig:qualitative}.}

\new{ChartCoder, while chart-aware, is trained on Chart2Code-160K---a fully synthetic dataset generated by GPT-4o using predefined attribute seeds with only Matplotlib and Seaborn as target libraries~\cite{zhao2025chartcoder}.
This limits its robustness on real-world charts with diverse styles, custom layouts, and non-standard visual encodings. As
illustrated in Fig.~\ref{fig:rose_case}, when reconstructing a customized pictorial infographic, ChartCoder cannot faithfully represent the rose-shaped glyphs using the predefined primitives provided by Matplotlib and instead approximates the petals with manually parameterized ellipses. Moreover, its executable Python-code output does not directly recover the original SVG structure or preserve its constituent graphical elements.}

\new{GPT-4o achieves competitive performance as a strong proprietary baseline but still falls short of Chart2SVG on most metrics, particularly in structural fidelity and failure rate, reflecting the benefit of chart-specific training and rendering-aware optimization.}

\new{\paragraph{Data Value Extraction Methodology.}
We evaluate data value accuracy (Table~3) using DIVI-based heuristic parsing of the rendered SVG rather than a VLM-based extraction oracle. We adopt this rule-based approach because numerical value extraction remains an open challenge across the field: even state-of-the-art VLMs such as o3 and Gemini-2.5-Flash achieve only moderate accuracy when 
reading data values directly from chart images, and dedicated chart-to-table models are typically restricted to charts with explicit, unambiguous value annotations or standard layouts, limiting their applicability to the diverse and customized charts in our evaluation. A rule-based parser operating directly on SVG geometry 
avoids the compounding uncertainty of VLM-based reading and provides a more reproducible, deterministic measurement of how faithfully mark geometry encodes the underlying data.}

\new{This limitation is not specific to Chart2SVG: no current chart-to-SVG or chart-to-code system embeds a dedicated, robust mechanism for precise numerical recovery, and we view this as an open problem for the broader field rather than a shortcoming unique to our 
method. We note, however, that the structured SVG representation produced by Chart2SVG offers a practical advantage over raster-only approaches in this regard: because mark geometry, axis ticks, and labels are recovered as separate, addressable elements rather than 
fused pixels, the reconstructed SVG can be readily combined with an external OCR module for tick-label reading, or with future versions of Chart2SVG that embed an OCR-aware decoding head, to achieve accurate axis-scale inference without retraining the core reconstruction model. We identify tighter integration of OCR-based 
numerical recovery, either as a post-hoc module or as an embedded component of the architecture, as a key direction for future work.}




%% file: chartsvg.bib
@article{poco2017reverse,
  title = {Reverse-Engineering Visualizations: Recovering Visual Encodings from Chart Images},
  author={Poco, Jorge and Heer, Jeffrey},
  journal = {Computer Graphics Forum},
  volume={36},
  number={3},
  pages={353-363},
  year={2017},
  organization={Wiley Online Library},
  doi = {10.1111/cgf.13193}
}

@inproceedings{savva2011revision,
  title={ReVision: Automated Classification, Analysis and Redesign of Chart Images},
  author={Savva, Manolis and Kong, Nicholas and Chhajta, Arti and Fei-Fei, Li and Agrawala, Maneesh and Heer, Jeffrey},
  booktitle={Proceedings of the 24th Annual ACM Symposium on User Interface Software and Technology},
  year={2011},
  doi = {10.1145/2047196.2047247}
}

@article{snyder2023divi,
  title={DIVI: Dynamically Interactive Visualization},
  author={Snyder, Luke S and Heer, Jeffrey},
  journal={IEEE Transactions on Visualization and Computer Graphics},
  volume={30},
  number={1},
  pages={403-413},
  year={2023},
  publisher={IEEE},
  doi={10.1109/TVCG.2023.3327172}
}

@inproceedings{rodriguez2025starvector,
  title={StarVector: Generating Scalable Vector Graphics Code from Images and Text},
  author={Rodriguez, Juan A and Puri, Abhay and Agarwal, Shubham and Laradji, Issam H and Rodriguez, Pau and Rajeswar, Sai and Vazquez, David and Pal, Christopher and Pedersoli, Marco},
  booktitle={2025 IEEE/CVF Conference on Computer Vision and Pattern Recognition (CVPR)}, 
  pages={16175-16186},
  year={2025},
  doi={10.1109/CVPR52734.2025.01508}
}

@inproceedings{xing2025empowering,
  title={Empowering LLMs to Understand and Generate Complex Vector Graphics}, 
  author={Xing, Ximing and Hu, Juncheng and Liang, Guotao and Zhang, Jing and Xu, Dong and Yu, Qian},
  booktitle={2025 IEEE/CVF Conference on Computer Vision and Pattern Recognition (CVPR)},
  pages={19487--19497},
  year={2025},
  doi={10.1109/CVPR52734.2025.01815}
}

@inproceedings{rodriguez2025rendering,
  title={Rendering-Aware Reinforcement Learning for Vector Graphics Generation},
  author={Rodriguez, Juan A and Zhang, Haotian and Puri, Abhay and Pramanik, Rishav and Feizi, Aarash and Wichmann, Pascal and Mondal, Arnab Kumar and Samsami, Mohammad Reza and Awal, Rabiul and Taslakian, Perouz and others},
  booktitle={The Thirty-ninth Annual Conference on Neural Information Processing Systems},
  year={2025}
}

@inproceedings{masson2024directgpt,
  title={DirectGPT: A Direct Manipulation Interface to Interact with Large Language Models},
  author={Masson, Damien and Malacria, Sylvain and Casiez, G{\'e}ry and Vogel, Daniel},
  booktitle={Proceedings of the 2024 CHI Conference on Human Factors in Computing Systems},
  pages={1-16},
  year={2024},
  doi = {10.1145/3613904.3642462},
}

@inproceedings{jung2017chartsense,
  title={ChartSense: Interactive Data Extraction from Chart Images},
  author={Jung, Daekyoung and Kim, Wonjae and Song, Hyunjoo and Hwang, Jeong-in and Lee, Bongshin and Kim, Bohyoung and Seo, Jinwook},
  booktitle = {Proceedings of the 2017 CHI Conference on Human Factors in Computing Systems},
  pages={6706-6717},
  year={2017},
  doi = {10.1145/3025453.3025957}
}

@article{bai2025qwen3,
  title={Qwen3-vl Technical Report},
  author={Bai, Shuai and Cai, Yuxuan and Chen, Ruizhe and Chen, Keqin and Chen, Xionghui and Cheng, Zesen and Deng, Lianghao and Ding, Wei and Gao, Chang and Ge, Chunjiang and others},
  journal={arXiv preprint arXiv:2511.21631},
  year={2025},
  doi={10.48550/arXiv.2511.21631}
}

@article{wang2025llm4dsr,
  title={Llm4dsr: Leveraging large language model for denoising sequential recommendation},
  author={Wang, Bohao and Liu, Feng and Zhang, Changwang and Chen, Jiawei and Wu, Yudi and Zhou, Sheng and Lou, Xingyu and Wang, Jun and Feng, Yan and Chen, Chun and others},
  journal={ACM Transactions on Information Systems},
  publisher={ACM New York, NY},
  doi={10.1145/3762182}
}

@article{shao2024deepseekmath,
  title={Deepseekmath: Pushing the Limits of Mathematical Reasoning in Open Language Models},
  author={Shao, Zhihong and Wang, Peiyi and Zhu, Qihao and Xu, Runxin and Song, Junxiao and Bi, Xiao and Zhang, Haowei and Zhang, Mingchuan and Li, YK and Wu, Yang and others},
  year={2024},
  doi={10.48550/arXiv.2402.03300}
}

@inproceedings{luo2021chartocr,
  title={ChartOCR: Data Extraction from Charts Images via a Deep Hybrid Framework},
  author={Luo, Junyu and Li, Zekun and Wang, Jinpeng and Lin, Chin-Yew},
  booktitle={2021 IEEE Winter Conference on Applications of Computer Vision (WACV)},
  pages={1916-1924},
  year={2021},
  doi={10.1109/WACV48630.2021.00196}
}

@inproceedings{hassan2023lineex,
  title={LineEX: Data Extraction from Scientific Line Charts},
  author={P, Shivasankaran V and Yusuf Hassan, Muhammad and Singh, Mayank},
  booktitle={2023 IEEE/CVF Winter Conference on Applications of Computer Vision (WACV)}, 
  pages={6202-6210},
  year={2023},
  doi={10.1109/WACV56688.2023.00615},
}

@inproceedings{cheng2023chartreader,
  title={ChartReader: A Unified Framework for Chart Derendering and Comprehension without Heuristic Rules}, 
  author={Cheng, Zhi-Qi and Dai, Qi and Hauptmann, Alexander G},
  booktitle={2023 IEEE/CVF International Conference on Computer Vision (ICCV)},
  pages={22145-22156},
  year={2023},
  doi={10.1109/ICCV51070.2023.02029}
}

@misc{selinger2003potrace,
  title={Potrace: a Polygon-Based Tracing Algorithm},
  author={Selinger, Peter},
  year={2003}
}

@article{li2020differentiable,
  title={Differentiable Vector Graphics Rasterization for Editing and Learning},
  author={Li, Tzu-Mao and Luk{\'a}{\v{c}}, Michal and Gharbi, Micha{\"e}l and Ragan-Kelley, Jonathan},
  journal={ACM Transactions on Graphics (TOG)},
  volume={39},
  number={6},
  pages={1-15},
  year={2020},
  publisher={Association for Computing Machinery},
  doi = {10.1145/3414685.3417871},
}

@inproceedings{carlier2020deepsvg,
  title={DeepSVG: A Hierarchical Generative Network for Vector Graphics Animation},
  author={Carlier, Alexandre and Danelljan, Martin and Alahi, Alexandre and Timofte, Radu},
  booktitle = {Advances in Neural Information Processing Systems},
  volume={33},
  year={2020}
}

@inproceedings{reddy2021im2vec,
  title={Im2Vec: Synthesizing Vector Graphics without Vector Supervision},
  author={Reddy, Pradyumna and Gharbi, Michael and Lukac, Michal and Mitra, Niloy J},
  booktitle={2021 IEEE/CVF Conference on Computer Vision and Pattern Recognition (CVPR)},
  pages={7338-7347},
  year={2021},
  doi={10.1109/CVPR46437.2021.00726}
}

@inproceedings{ma2022towards,
  title={Towards Layer-wise Image Vectorization},
  author={Ma, Xu and Zhou, Yuqian and Xu, Xingqian and Sun, Bin and Filev, Valerii and Orlov, Nikita and Fu, Yun and Shi, Humphrey},
  booktitle={2022 IEEE/CVF Conference on Computer Vision and Pattern Recognition (CVPR)},
  pages={16293-16302},
  year={2022},
  doi={10.1109/CVPR52688.2022.01583}
}

@article{chen2025visanatomy,
  title={VisAnatomy: An SVG Chart Corpus with Fine-Grained Semantic Labels}, 
  year={2026},
  volume={32},
  number={1},
  pages={560-570},
  author={Chen, Chen and Bako, Hannah K and Yu, Peihong and Hooker, John and Joyal, Jeffrey and Wang, Simon C and Kim, Samuel and Wu, Jessica and Ding, Aoxue and Sandeep, Lara and others},
  journal={IEEE Transactions on Visualization and Computer Graphics},
  publisher={IEEE},
  doi={10.1109/TVCG.2025.3634263}
}

@inproceedings{mendez2016ivolver,
  title={iVoLVER: Interactive Visual Language for Visualization Extraction and Reconstruction},
  author={M{\'e}ndez, Gonzalo Gabriel and Nacenta, Miguel A and Vandenheste, Sebastien},
  booktitle={Proceedings of the 2016 CHI Conference on Human Factors in Computing Systems},
  pages={4073-4085},
  year={2016},
  doi={10.1145/2858036.2858435}
}

@article{bako2022understanding,
  author={Bako, Hannah K. and Liu, Xinyi and Battle, Leilani and Liu, Zhicheng},
  journal={IEEE Transactions on Visualization and Computer Graphics}, 
  title={Understanding how Designers Find and Use Data Visualization Examples}, 
  year={2023},
  volume={29},
  number={1},
  pages={1048-1058},
  doi={10.1109/TVCG.2022.3209490}
}

@article{bigelow2016iterating,
  author={Bigelow, Alex and Drucker, Steven and Fisher, Danyel and Meyer, Miriah},
  journal={IEEE Transactions on Visualization and Computer Graphics}, 
  title={Iterating between Tools to Create and Edit Visualizations}, 
  year={2017},
  doi={10.1109/TVCG.2016.2598609}
}

@article{yin2024survey,
  title={A Survey on Multimodal Large Language Models},
  author={Yin, Shukang and Fu, Chaoyou and Zhao, Sirui and Li, Ke and Sun, Xing and Xu, Tong and Chen, Enhong},
  journal={National Science Review},
  volume={11},
  number={12},
  pages={nwae403},
  year={2024},
  publisher={Oxford University Press},
  doi={10.1093/nsr/nwae403}
}

@book{harder2023creating,
  title={Creating Infographics with Adobe Illustrator: Volume 2},
  author={Harder, Jennifer}
}

@article{chen2023mystique,
  author={Chen, Chen and Lee, Bongshin and Wang, Yunhai and Chang, Yunjeong and Liu, Zhicheng},
  journal={IEEE Transactions on Visualization and Computer Graphics}, 
  title={Mystique: Deconstructing SVG Charts for Layout Reuse}, 
  year={2024},
  volume={30},
  number={1},
  pages={447-457},
  doi={10.1109/TVCG.2023.3327354}
}

@article{cui2021mixed,
   author={Cui, Weiwei and Wang, Jinpeng and Huang, He and Wang, Yun and Lin, Chin-Yew and Zhang, Haidong and Zhang, Dongmei},
  journal={IEEE Transactions on Visualization and Computer Graphics}, 
  title={A Mixed-Initiative Approach to Reusing Infographic Charts}, 
  year={2022},
  volume={28},
  number={1},
  pages={173-183},
  doi={10.1109/TVCG.2021.3114856}
}

@inproceedings{chen2024onechart,
  title={OneChart: Purify the Chart Structural Extraction via One Auxiliary Token},
  author={Chen, Jinyue and Kong, Lingyu and Wei, Haoran and Liu, Chenglong and Ge, Zheng and Zhao, Liang and Sun, Jianjian and Han, Chunrui and Zhang, Xiangyu},
  booktitle={Proceedings of the 32nd ACM International Conference on Multimedia},
  pages={147-155},
  year={2024},
  doi={10.1145/3664647.3681167},
}

@inproceedings{yangomnisvg,
  title={OmniSVG: A Unified Scalable Vector Graphics Generation Model},
  author={Yang, Yiying and Cheng, Wei and Chen, Sijin and Zeng, Xianfang and Yin, Fukun and Zhang, Jiaxu and Wang, Liao and YU, Gang and Ma, Xingjun and Jiang, Yu-Gang},
  booktitle={The Thirty-ninth Annual Conference on Neural Information Processing Systems}
}

@inproceedings{zhao2025chartcoder,
  title={Chartcoder: Advancing Multimodal Large Language Model for Chart-to-Code Generation},
  author={Zhao, Xuanle and Luo, Xianzhen and Shi, Qi and Chen, Chi and Wang, Shuo and Liu, Zhiyuan and Sun, Maosong},
  booktitle={Proceedings of the 63rd Annual Meeting of the Association for Computational Linguistics (Volume 1: Long Papers)},
  pages={7333-7348},
  year={2025},
  doi={10.18653/v1/2025.acl-long.363},
}

@article{choi2019visualizing,
  title={Visualizing for the Non-Visual: Enabling the Visually Impaired to Use Visualization},
  author={Choi, Jinho and Jung, Sanghun and Park, Deok Gun and Choo, Jaegul and Elmqvist, Niklas},
  journal={Computer graphics forum},
  volume={38},
  number={3},
  pages={249-260},
  year={2019},
  organization={Wiley Online Library},
  doi={10.1111/cgf.13686}
}

@inproceedings{battle2018beagle,
  title={Beagle: Automated Extraction and Interpretation of Visualizations from the Web},
  author={Battle, Leilani and Duan, Peitong and Miranda, Zachery and Mukusheva, Dana and Chang, Remco and Stonebraker, Michael},
  booktitle={Proceedings of the 2018 CHI Conference on Human Factors in Computing Systems},
  doi={10.1145/3173574.3174168},
}

@article{li2018echarts,
  title={ECharts: a Declarative Framework for Rapid Construction of Web-based Visualization},
  author={Li, Deqing and Mei, Honghui and Shen, Yi and Su, Shuang and Zhang, Wenli and Wang, Junting and Zu, Ming and Chen, Wei},
  journal={Visual Informatics},
  volume={2},
  number={2},
  pages={136-146},
  year={2018},
  publisher={Elsevier},
  doi={10.1016/j.visinf.2018.04.011}
}

@article{harper2017converting,
  author={Harper, Jonathan and Agrawala, Maneesh},
  journal={IEEE Transactions on Visualization and Computer Graphics}, 
  title={Converting Basic D3 Charts into Reusable Style Templates}, 
  year={2018},
  volume={24},
  number={3},
  pages={1274-1286},
  doi={10.1109/TVCG.2017.2659744}
}

@inproceedings{mcnutt2021integrated,
  title={Integrated Visualization Editing via Parameterized Declarative Templates},
  author={McNutt, Andrew M and Chugh, Ravi},
  booktitle={Proceedings of the 2021 CHI Conference on Human Factors in Computing Systems},
  pages={1-14},
  year={2021},
  doi={10.1145/3411764.3445356}
}

@inproceedings{wu2025plot2code,
  title={Plot2code: A Comprehensive Benchmark for Evaluating Multi-modal Large Language Models in Code Generation from Scientific Plots},
  author={Wu, Chengyue and Liang, Zhixuan and Ge, Yixiao and Guo, Qiushan and Lu, Zeyu and Wang, Jiahao and Shan, Ying and Luo, Ping},
  booktitle={Findings of the Association for Computational Linguistics: NAACL 2025},
  pages={3006-3028},
  year={2025},
  doi={10.18653/v1/2025.findings-naacl.164}
}

@article{xia2025chartx,
  author={Xia, Renqiu and Ye, Hancheng and Yan, Xiangchao and Liu, Qi and Zhou, Hongbin and Chen, Zijun and Shi, Botian and Yan, Junchi and Zhang, Bo},
  journal={IEEE Transactions on Image Processing}, 
  title={ChartX and ChartVLM: A Versatile Benchmark and Foundation Model for Complicated Chart Reasoning}, 
  year={2025},
  volume={34},
  number={},
  pages={7436-7447},
  doi={10.1109/TIP.2025.3607618}
}

@inproceedings{hu2022lora,
  title={Lora: Low-Rank Adaptation of Large Language Models.},
  author={Hu, Edward J and Shen, Yelong and Wallis, Phillip and Allen-Zhu, Zeyuan and Li, Yuanzhi and Wang, Shean and Wang, Liang and Chen, Weizhu and others},
  booktitle={International Conference on Learning Representations (ICLR)},
  year={2022},
  url={https://openreview.net/forum?id=nZeVKeeFYf9}
}

@misc{svgo,
  key={svgo},
  authors={SVG Contributors},
  title={SVGO: Node.js tool for optimizing SVG files},
  howpublished = {\url{ https:
//github.com/svg/svgo}},
  note = {Accessed: 2026-03-04}
}

@article{xie2025datawink,
  author={Xie, Liwenhan and Lin, Yanna and Liu, Can and Qu, Huamin and Shu, Xinhuan},
  journal={IEEE Transactions on Visualization and Computer Graphics}, 
  title={DataWink: Reusing and Adapting SVG-Based Visualization Examples with Large Multimodal Models}, 
  year={2026},
  volume={32},
  number={1},
  pages={824-834},
  doi={10.1109/TVCG.2025.3634635}
}

@inproceedings{wang2026msrl,
  title={MSRL: Scaling Generative Multimodal Reward Modeling via Multi-Stage Reinforcement Learning},
  author={Wang, Chenglong and Huo, Yifu and Gan, Yang and He, Qiaozhi and Meng, Qi and Li, Bei and Wang, Yan and Liu, Junfu and Zhou, Tianhua and Zhu, Jingbo and others},
  booktitle={Proceedings of the IEEE/CVF Conference on Computer Vision and Pattern Recognition},
  pages={29410--29420},
  year={2026}
}

@article{xing2025reason,
  title={Reason-svg: Hybrid reward rl for aha-moments in vector graphics generation},
  author={Xing, Ximing and Guan, Yandong and Zhang, Jing and Xu, Dong and Yu, Qian},
  journal={arXiv preprint arXiv:2505.24499},
  volume={4},
  year={2025}
}
